\PassOptionsToPackage{expansion=false}{microtype}
\documentclass[manuscript,screen,nonacm]{acmart}

\AtBeginDocument{%
  }

\setcopyright{none}
\acmYear{2027}

\usepackage{booktabs}
\usepackage{colortbl}
\usepackage[most]{tcolorbox}
\newtcolorbox{promptbox}[1]{breakable, enhanced, colback=blue!4, colframe=blue!45!black, boxrule=0.6pt, arc=2pt, left=6pt, right=6pt, top=4pt, bottom=4pt, fonttitle=\bfseries\small, title={#1}, fontupper=\footnotesize}
\usepackage{algorithm}
\usepackage{algpseudocode}
\usepackage{tikz}
\usetikzlibrary{positioning,arrows.meta,fit}

\newcommand{\KH}{\textsc{KnifeHunter}}
\newcommand{\BWVP}{UKBWV-27}

\begin{document}

\title{From Weapons to Events: Evaluating Evidential Support in Multimodal AI for Policing}

\author{Armin Mustafa}
\affiliation{%
  \institution{Centre for Vision, Speech and Signal Processing, University of Surrey}
  \city{Guildford}
  \country{United Kingdom}}
\affiliation{%
  \institution{AI Fellow, Incubator for AI (i.AI), Department for Digital, Culture, Media and Sport}
  \city{London}
  \country{United Kingdom}}
\email{armin.mustafa@surrey.ac.uk}

\author{Syed Sameed Husain}
\affiliation{%
  \institution{Centre for Vision, Speech and Signal Processing, University of Surrey}
  \city{Guildford}
  \country{United Kingdom}}

\renewcommand{\shortauthors}{Mustafa and Husain}

\begin{abstract}
Computer vision (CV) for policing has centred on recognising objects such as seized weapons, while body-worn video (BWV) is still reviewed by people, with existing AI tools using only its audio. Multimodal video AI could make BWV review faster and cheaper.
At the object level, we present the KnifeHunter dataset, developed with UK law enforcement, with over 25,000 images of 543 knife classes and a protocol reflecting investigative conditions, and KHNet, a retrieval architecture fusing local and global features, which with about 125,000 distractor images attains $85.7\%$ and $63.4\%$ mean Average Precision (mAP) on the medium and hard protocols.
At the event level, we present an evaluation protocol for video understanding pipelines, which describe a video as events and a transcript, that scores transcription, speaker attribution, event detection, temporal localisation and unsupported events. Whether each event is supported by the recording is labelled by a judge validated against blinded human labels. We apply it to complete pipelines, three 2026 omni-modal models and our hybrid NarrativeBridge--PAL--GPT-5 pipeline, on UKBWV-27, a pilot stress-test set of 27 publicly released UK BWV recordings (118 minutes). On this set the hybrid pipeline leads on five of six measures, yet it leaves 13.6\% of predicted events unsupported and misattributes 27.7\% of utterances, and every pipeline degrades on recordings from edited releases, long recordings and noisy audio. We propose readiness levels separating assistive tasks from those requiring verification of every output.
\end{abstract}



\maketitle

\section{Introduction}
\label{sec:intro}

\subsection{Background and motivation}
Body-worn video (BWV) cameras were adopted by police services in the United Kingdom and elsewhere with the stated aims of improving transparency, gathering evidence and supporting the accountability of officers to the public. Randomised and quasi-experimental studies have examined their effects on use of force and complaints \cite{ariel2015bwv}, and a systematic review found that effects on most officer and citizen behaviours are neither clear nor consistent \cite{lum2020bwv}. A precondition for any accountability benefit is that recorded footage is actually reviewed. The volume is substantial: in London's year-long randomised trial, 814 officers made 48,281 recordings totalling 12,212 hours, and officers reported that prosecutors and secondary investigators were often unable to view footage; the Metropolitan Police then announced plans to buy around 20,000 cameras \cite{grossmith2015bwv}. Manual review therefore cannot keep pace, so in practice review is selective and driven by complaints or incidents rather than by systematic sampling.
Computational analysis can measure patterns in BWV at a scale manual review cannot reach \cite{voigt2017respect}.

AI in policing is not limited to BWV. Knife-enabled violence is a persistent public safety problem in the UK, and officers need to identify recovered or seized knives against catalogues of known products so that they can trace sources and analyse trends, yet recovered knives are often stored with only a brief text description, which prevents systematic search. This is an object-level recognition problem with a well-defined ground truth (the catalogue identity of a knife), in contrast to the event-level problem of describing what happened in a BWV recording, where the ground truth is a structured interpretation of a complex interaction.

\subsection{Research gap}
Video AI has advanced rapidly, but the evidence for its use in policing is thin. We identify three gaps. (i) \emph{No evaluation of video AI on BWV}: the AI tools studied in police BWV workflows work only from the audio track, analysing transcripts \cite{white2025truleo} or drafting reports \cite{adams2024nomanshand}; to our knowledge, no published work evaluates models that analyse the video itself. (ii) \emph{Evaluation without correctness}: those tools were judged on adoption and time saved, general video benchmarks test question answering \cite{fu2025videomme}, and accountability frameworks call for audits and documentation \cite{raji2020closing,mitchell2019modelcards,gebru2021datasheets} without specifying task metrics. None checks, for police footage, whether each reported event actually appears in the recording at the right time, or whether words are attributed to the right speaker. (iii) \emph{Fragmented evidence}: to our knowledge, no prior work evaluates object-level and event-level police AI systems side by side.
\vspace{-0.3cm}
\subsection{Argument}
\label{sec:argument}
The use of CV for police tasks has so far centred on recognising objects in images, such as identifying seized weapons with \KH{}, while BWV is still reviewed by people watching it; the AI tools used on it so far work only from its audio \cite{white2025truleo,adams2024nomanshand}. We argue that \emph{advances in video AI could make CV a realistic aid to BWV review that reduces its cost and time}, and give pilot evidence on \BWVP{}: the best pipeline detects events with an F1 of 63.1, and every pipeline processes a minute of footage in under 39 seconds, faster than watching it. Because police records carry consequences, such use would be acceptable only if \emph{each AI-generated claim can be traced to the footage, its failures are visible, and people can contest it}. We call these conditions \emph{traceability}, \emph{failure visibility} and \emph{contestability}. They are design goals, not results: we measure properties of the outputs that each condition needs, not whether people can actually trace, spot or contest errors. For traceability, we measure how often events are unsupported by the footage, using a judge checked against blinded human labels, and how well events are located in time. For failure visibility, we report results per recording condition, so that failures are not hidden in one average. For contestability, we measure speaker attribution, because it decides whose words a record contains; whether people can contest those records is not measured. \KH{} is the contrasting case: its ground truth is objective and a person confirms every output.
\vspace{-0.3cm}
\subsection{Research questions and contributions}
We address the following research questions.
\begin{description}
\item[RQ1] How can multimodal AI for policing be evaluated for traceability, failure visibility and contestability?
\item[RQ2] How accurately can retrieval architectures identify knives under operational conditions, and how does accuracy change with query difficulty and gallery size?
\item[RQ3] How well do current multimodal pipelines perform event-level BWV review tasks on a pilot set, and what are their characteristic failure modes?
\end{description}
Our contributions answer these questions in turn.
\begin{enumerate}
\item \textbf{A pilot benchmark, \BWVP{} and an evaluation protocol} (RQ1): a protocol that checks whether each reported event can be seen or heard in the BWV, separately from accuracy, using an automated judge validated against blinded human labels; and 27 public UK BWV recordings (118 minutes after removing news inserts and overlays) with machine-assisted, human-corrected annotations (Sections~\ref{sec:bwv-data} to \ref{sec:judge}).
\item \textbf{The \KH{} dataset and KHNet} (RQ2): 25,250 images of 543 catalogued knife classes with structured metadata, built with UK law enforcement; a protocol with medium and hard queries and 125,000 distractors; and KHNet, a retrieval network that fuses regionally pooled local features with Weibull-activated global features, deployed in UK policing (Sections~\ref{sec:caseA} and \ref{sec:res-object}).
\item \textbf{A pipeline benchmark on BWV} (RQ3): our hybrid NarrativeBridge--PAL--GPT-5 pipeline against three 2026 omni-modal models (Qwen3.5-Omni-Plus \cite{qwen2026qwen35omni}, Nemotron 3 Nano Omni \cite{nvidia2026nemotronomni} and Gemini 3.1 Pro \cite{google2026gemini31}), compared as complete pipelines, and the failure modes this reveals (Section~\ref{sec:benchmark}).
\end{enumerate}

\section{Related Work}
\label{sec:related}

\subsection{Body-worn video, AI and police accountability}
BWV was introduced as an accountability technology, with mixed evidence: a randomised trial found use of force roughly half as likely on shifts with cameras \cite{ariel2015bwv}, London's trial found 33\% fewer allegations per officer in camera-wearing teams \cite{grossmith2015bwv}, and a systematic review found effects inconsistent across studies \cite{lum2020bwv}. Footage has also been analysed as data, for example to measure officer respect in traffic stops \cite{voigt2017respect}. AI tools are now entering BWV workflows faster than they are evaluated: a system analysing BWV audio in two US departments was used routinely without its accuracy being assessed \cite{white2025truleo}, and a randomised trial of a tool drafting reports from BWV audio found no change in completion time \cite{adams2024nomanshand}. Because data-driven policing can encode and amplify existing biases \cite{ferguson2017bigdata,richardson2019dirty}, we treat automated BWV understanding as an input to human review whose outputs must be evaluated under the conditions in which it is used.
\vspace{-0.2cm}
\subsection{Fine-grained visual retrieval}
Identifying a knife against a catalogue is fine-grained instance retrieval: classes differ in details of blade, serration and handle, some of which determine legal classification. Instance retrieval has progressed from aggregating local descriptors \cite{husain2017rvd,husain2019remap} to end-to-end global descriptors learned with generalized-mean pooling \cite{radenovic2019gem}, learnable Weibull activations \cite{husain2021actnet}, token-based attention \cite{wu2022tokennet}, self-similarity \cite{lee2023senet} and orthogonal local--global fusion \cite{yang2021dolg}, evaluated by mAP \cite{radenovic2018revisiting}. No public benchmark existed for knives; we build one and adapt these aggregation methods to it.
\vspace{-0.2cm}
\subsection{Multimodal video understanding and its evaluation}
Dense video captioning localises and describes multiple events in untrimmed video \cite{krishna2017dense}. Causal-temporal approaches extend captioning to describe why events occur and in what order: NarrativeBridge introduced a causal-temporal narrative benchmark and a cause-effect network for this purpose \cite{nadeem2025narrativebridge}, and Causal-VLM addresses dense causal captioning \cite{nadeem2026causalvlm}. Audio is a necessary modality for BWV, where speech carries much of the meaning of an interaction; large-scale speech recognition \cite{radford2023whisper} and methods for transferring audio encoder information into language models \cite{alex2025pal} provide components for audio-aware video understanding, and recent omni-modal models accept audio and video jointly \cite{qwen2026qwen35omni,nvidia2026nemotronomni,google2026gemini31}. Evaluation has moved from n-gram and embedding metrics \cite{vedantam2015cider,zhang2020bertscore} to video question-answering benchmarks \cite{fu2025videomme} and language-model judges \cite{zheng2023judging}, which have systematic biases \cite{wang2024fair}; we therefore validate our judge against blinded human labels.
\vspace{-0.2cm}
\subsection{Evaluation, accountability and fairness}
Accountability requires end-to-end auditing \cite{raji2020closing}, documentation of models and data \cite{mitchell2019modelcards,gebru2021datasheets} and attention to the context of use \cite{selbst2019abstraction}; we follow this with a datasheet and per-task, per-condition results. People over-rely on automated aids \cite{skitka1999automation} and do not use algorithmic advice neutrally \cite{green2019disparate}, which is why we treat human review as a condition to be tested, not assumed. Disaggregated evaluation has revealed disparities in face analysis \cite{buolamwini2018gender} and speech recognition \cite{koenecke2020asr}, the latter directly relevant to BWV transcription. Evaluation measures should also be validated with stakeholders \cite{nguyen2026validating}, and speech recognition audits have pitfalls of their own \cite{mei2026aphasia}.

\section{Methodology}
\label{sec:method}
\subsection{Pipelines and tasks}
\label{sec:taxonomy}
We evaluate four police AI tasks and present a deployed knife retrieval system alongside an experimental BWV pipeline, which has not been used by police and serves only to show whether such analysis could be useful for policing, and label claims by evidence type: \emph{E1}, measurements and the deployment record of \KH{} (Case A); \emph{E2}, new measurements on \BWVP{} (Case B); and \emph{E3}, analytical argument (cross-case comparison, applications and readiness).

\paragraph{Object level (Case A)} A photograph of a recovered or seized knife is ranked against a product catalogue, and the ranked matches with their metadata are the output. \emph{T1}: identifying the knife, scored by mAP.

\begin{figure*}[t]
\centering
\resizebox{\textwidth}{!}{%
\begin{tikzpicture}[
  every node/.append style={execute at begin node={\hyphenpenalty=10000\relax}},
  stage/.style={draw, rounded corners=2pt, align=center, font=\small, minimum height=2.15cm, text width=2.9cm, fill=white},
  fixed/.style={stage, fill=gray!12},
  hum/.style={stage, dashed, fill=gray!12},
  task/.style={draw, rounded corners=2pt, align=center, font=\footnotesize, minimum height=1.3cm, text width=3.0cm, fill=blue!6},
  arr/.style={-{Stealth[length=2mm]}, thick},
  darr/.style={-{Stealth[length=2mm]}, thick, dashed},
  lab/.style={font=\scriptsize\itshape, align=center},
  node distance=0.6cm
]
\node[stage] (in) {\textbf{1. Input}\\[2pt]BWV recording,\\non-BWV parts removed\\(frames and audio)};
\node[fixed, right=of in] (ch) {\textbf{2. Chapterisation}\\[2pt]ARC-Chapter, run once;\\same fixed chapters\\for every system};
\node[stage, right=of ch] (gen) {\textbf{3. Generation}\\[2pt]base system, per chapter:\\event claims (0 or more),\\with times and roles;\\transcript};
\node[stage, right=of gen] (pp) {\textbf{4. Post-processing}\\[2pt]GPT-5, text only:\\event labels, roles,\\utterance links};
\node[hum, right=of pp] (rev) {\textbf{5. Human review}\\[2pt]check each claim\\against the footage};
\draw[arr] (in) -- (ch);
\draw[arr] (ch) -- (gen);
\draw[arr] (gen) -- (pp);
\draw[darr] (pp) -- (rev);
\node[lab, above=0.1cm of ch] {fixed; not evaluated};
\node[lab, above=0.1cm of gen] {compared systems};
\node[lab, above=0.1cm of pp] {same prompt for all};
\node[lab, above=0.1cm of rev] {proposed; not evaluated};
\node[task, below=1.5cm of gen, xshift=-0.3cm] (t2) {\textbf{T2} Detect and locate events\\[1pt]event F1, mIoU};
\node[task, right=0.35cm of t2] (t3) {\textbf{T3} Unsupported claims\\[1pt]$U$, labelled by a validated judge};
\node[task, right=0.35cm of t3] (t4) {\textbf{T4} Transcript and speaker\\[1pt]WER, speaker accuracy};
\node[draw, dotted, rounded corners=3pt, fit=(t2)(t3)(t4), inner sep=6pt] (box) {};
\node[lab, above right=0.05cm and 0.1cm of box.north west, align=left] {scored against the \BWVP{} reference annotations;\\every claim keeps its time interval};
\draw[arr] (pp.south) -- (pp.south |- box.north);
\draw[darr] (in.south) -- ([yshift=-0.75cm]in.south |- box.south) -| node[pos=0.25, above, lab]{recording also given to the judge} (t3.south);
\end{tikzpicture}}
\caption{Event-level BWV pipeline and where each task is scored. A fixed chapteriser divides each recording once into chapters shared by all systems. Each base system generates event claims and a transcript within those chapters, and GPT-5 maps them to the common schema without adding, removing or retiming events. Tasks T2 to T4 are scored on the post-processed output. Shaded boxes are not evaluated; human review is proposed, not tested.}
\Description{A row of five boxes from left to right: input recording with non-BWV parts removed; chapterisation by ARC-Chapter, run once and shared by every system; generation by the compared base system, producing event claims and a transcript within each chapter; post-processing by GPT-5, producing event labels, roles and utterance links; and a dashed box for proposed human review. An arrow from post-processing leads down to three task boxes: T2 event detection scored by event F1 and mIoU, T3 unsupported claims scored by the unsupported-event rate via a validated judge, and T4 transcript and speaker attribution scored by WER and speaker accuracy. A dashed line shows that the recording is also given to the judge.}
\label{fig:bwv-pipeline}
\vspace{-0.5cm}
\end{figure*}
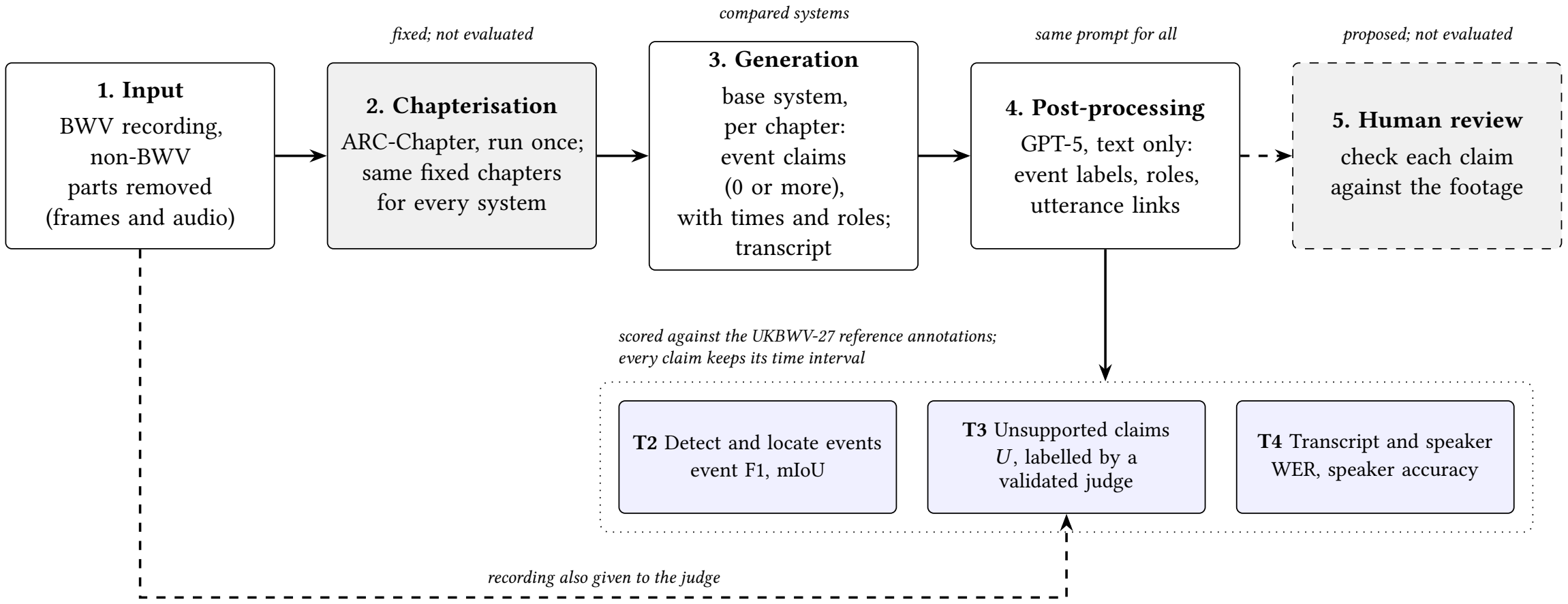

\paragraph{Event level (Case B)} Events are described with a fixed list of 16 event types, police actions and interactions such as Taser use, handcuffing and restraint, and roadside stop and search (full list in Table~\ref{tab:inventory}, Appendix~\ref{app:guidelines}). The same list is used to label the reference annotations, is given to every system, and is the only set of labels that is scored. A BWV recording passes through five stages (Figure~\ref{fig:bwv-pipeline}).
\begin{enumerate}
\item \emph{Input.} Frames and the audio track of the recording.
\item \emph{Chapterisation.} A fixed model divides the recording once into chapters, which give every system the same context intervals and are not evaluated.
\item \emph{Generation.} In each chapter, a base system lists zero or more events with their start and end times. For each event, it suggests one of the 16 event types and says, by role only, who did it and to whom (for example, an officer restraining a member of the public); systems that output only free-text descriptions, such as NarrativeBridge in our hybrid pipeline, leave these to stage~4. It also transcribes the speech, labelling each line as said by an officer, a member of the public or someone else.
\item \emph{Post-processing.} GPT-5 maps each event to one of 16 event types or to unmapped, settles actor and target roles and links utterances to events, without adding, removing or retiming events.
\item \emph{Human review.} A reviewer would check the outputs against the footage; this stage is proposed (Section~\ref{sec:applications}).
\end{enumerate}
The outputs of stage~4 are scored as three tasks, which were identified from meetings with two staff members at the partner police service, who explained how BWV is currently reviewed, what a review tool would have to do and where video understanding could help.
\emph{T2}: detecting and locating events to triage videos, scored by event F1 and mIoU. \emph{T3}: reporting events without unsupported claims, scored by the unsupported-event rate. \emph{T4}: transcribing speech and attributing it to the right speaker, scored by WER and speaker accuracy. Each task's output carries a time interval, so a reviewer can go from any claim to the part of BWV that should support it.
\vspace{-0.3cm}
\subsection{Case A: fine-grained knife retrieval (\KH{})}
\label{sec:caseA}
\paragraph{Dataset.} We built the \KH{} dataset with a UK police service. Its database contains 25,250 images of 543 catalogued knife classes from three sources: 332 classes photographed in a police evidence store with phones and DSLR cameras under varied viewpoints, lighting and backgrounds; 171 classes from a UK retailer's product images, with augmentation; and 40 classes seized by UK Border Force. Each class carries ten metadata fields developed with police forces, manufacturers and retailers, including knife type, manufacturer, product name, a flag for whether it could be an offensive weapon under UK law, blade length, material and stockist.

\paragraph{Evaluation protocol.} The Medium set has 443 query images with moderate viewpoint change, partial occlusion, illumination change and clutter, some from knife-crime operations. The Hard set has 150 query images, many from knife-crime investigations, with unusual angles, heavy occlusion, clutter, extreme lighting or inverted knives. A distractor set of about 125,000 images, combining knife images supplied by nine UK police forces (none of them in the dataset) with unlabelled COCO images \cite{lin2014coco}, can be added to the database (the +D protocols). Each query is ranked against the database by cosine similarity of global descriptors. With $\mathrm{rel}_q(i)\in\{0,1\}$ indicating whether the item at rank $i$ shares the query's class, $R_q$ the number of relevant items and $P_q(i)$ the precision of the top $i$ items, we report mean average precision \cite{radenovic2018revisiting}:
\begin{equation}
\mathrm{AP}(q)=\frac{1}{R_q}\sum_{i=1}^{|\mathcal{G}|}P_q(i)\,\mathrm{rel}_q(i),\qquad \mathrm{mAP}=\frac{1}{|Q|}\sum_{q\in Q}\mathrm{AP}(q).
\label{eq:ap}
\end{equation}

\paragraph{Architectures.} Each model is a backbone (EfficientNet-B7 \cite{tan2019efficientnet}, ConvNeXt-Base \cite{liu2022convnext}, SwinV2-Base \cite{liu2022swinv2} or EVA02-Base \cite{fang2024eva02}) followed by an aggregation head: GeMNet \cite{radenovic2019gem}, ActNet \cite{husain2021actnet}, TokenNet \cite{wu2022tokennet}, SENet \cite{lee2023senet} or our KHNet, which fuses regionally pooled local features with Weibull-activated global features by orthogonal projection \cite{yang2021dolg} (Appendix~\ref{app:kh-ablation}). All models are trained end to end with sub-center ArcFace \cite{deng2020subcenter}.

\paragraph{Deployment.} \KH{} runs on a cloud CPU node that processes up to 50 images per second. Officers upload a photograph through a web page and receive an automatically generated report of the retrieved knife details. It was used in two rounds of a national knife-crime campaign and by forces across the UK. Its deployment record documents use, not field accuracy, which has not been measured.
\vspace{-0.3cm}
\subsection{Case B: experimental multimodal BWV understanding pipeline}
\label{sec:caseB}
\paragraph{Formal definition.} A BWV recording $V$ is represented by sampled frames, the audio signal and metadata. The pipeline of Section~\ref{sec:taxonomy} produces an output set $\mathcal{O}(V)=\{\mathcal{E},W\}$, where $\mathcal{E}$ is the set of predicted events $e=(\ell,[t_s,t_e])$ pooled over all chapters, each with label $\ell$ and an interval inside its chapter; and $W$ is a transcript whose utterances carry a speaker role and timestamps. Every element carries a time interval, which we call its \emph{evidence anchor}. A chapter is a shared context interval, not a predicted event: the event is the unit of evaluation for detection, temporal overlap and support, and chapters themselves are not scored.

\paragraph{Common chapterisation.} The chapteriser is ARC-Chapter \cite{pu2025arcchapter} (on Qwen2.5-VL-7B), run once per recording on its frames and audio and frozen before evaluation; its chapter identifiers and timestamps are supplied unchanged to every base system, and it is not part of the model comparison (Appendix~\ref{app:bench-protocol}). Models and inference are specified in Section~\ref{sec:benchmark}. Faces are not redacted before model input because all footage is already public.

\subsubsection{Pilot benchmark: \BWVP{}}
\label{sec:bwv-data}
\paragraph{Scope.} \BWVP{} is a pilot stress-test set, not a representative sample of UK BWV, and its results do not show operational reliability. Drawn from public releases, it over-represents high-profile, severe and edited releases and under-represents routine encounters, and publisher editing may affect performance.

\paragraph{Sources and selection.} \BWVP{} contains 27 recordings from 16 UK police forces, totalling 118 minutes (26\,s to 15\,min 44\,s; median 2\,min 26\,s), published between 2017 and 2026: 3 by police forces and 24 by news outlets, including footage shown in court, at an inquest and at a misconduct hearing. All 27 are used for benchmarking; there is no training split, because no model is trained or fine-tuned on BWV.
Selection criteria are: (i) consisting wholly or mainly of body-worn camera footage; (ii) published by a police force or a news outlet; (iii) containing at least one interaction between an officer and a member of the public; and (iv) covering variation in lighting, camera motion, occlusion, audio quality and number of people. Sources, release terms and jurisdictions are recorded per video in the datasheet (Appendix~\ref{app:datasheet}) \cite{gebru2021datasheets}. Summary statistics are given in Appendix~\ref{app:datasheet} (Table~\ref{tab:bwv-stats}).

\paragraph{Preparation.} Because most recordings were republished by news outlets, we removed, before evaluation, every segment not filmed by a body-worn camera (such as studio, CCTV or helicopter inserts) and every segment carrying broadcaster overlays such as logos or captions, so that models received only clean BWV. Segments were identified from the recording's visual source and editing properties only, not from the incident, events or headline. All durations above refer to the retained footage. Recordings are still grouped by how they were released (22 as pure BWV, 3 with overlays, 2 combined with other sources), because publisher editing can leave cuts and compression in the retained footage. The recordings and news coverage of these incidents are public, so hosted models may have seen them in training, and results may not reflect perception of the footage alone.

\paragraph{Annotation protocol.} Each video is annotated with a speech transcript, with speaker roles (officer, member of the public, other or unknown) and timestamps, and with events, each an interval labelled with one of 16 event types (Appendix~\ref{app:guidelines}, Table~\ref{tab:inventory}), plus an unscored video-level tag, taken from the publication context, for force later challenged in court or at a hearing or inquest. Events are annotated only if visible or clearly audible in body-worn footage, last at least 1\,s and may overlap; inclusion rules and borderline cases are in Appendix~\ref{app:guidelines}.  Two open omni-modal models, Baichuan-Omni-1.5 \cite{li2025baichuanomni15} and MiMo-V2.5 \cite{xiaomi2026mimov25}, generated initial event annotations, which two human annotators then reviewed and corrected against the video and audio. These are machine-assisted, human-corrected reference annotations, not gold-standard labels. Correcting model proposals can anchor annotators, omit events the models did not propose, and carry over model boundaries and interpretations, especially for force, restraint and actor and target roles. Annotators did not first annotate the raw footage independently, agreement before adjudication was not measured, and the number of events added, removed or changed after pre-annotation is not reported. The reference set is complete only with respect to the 16 types. Both pre-annotation models are excluded from the evaluated panel. Transcripts were initialised with Whisper \cite{radford2023whisper} and then corrected by the same annotators against the audio.
Annotators do not label perceived race, ethnicity, gender or age of people in the footage.

\subsubsection{Metrics for event-level tasks}
\label{sec:metrics}
Let $\mathrm{tIoU}(u,v)=|u\cap v|/|u\cup v|$ denote the temporal intersection over union of two intervals.

\paragraph{Event detection and temporal localisation (T2).} Event-level metrics are computed over the union of predicted events produced within all fixed chapters. A chapter may contain zero, one, or multiple predicted events. Predicted events from different chapters are pooled and matched one-to-one with human reference events using Hungarian matching, temporal IoU $\geq 0.3$, and equivalent event labels. Events may overlap or nest, including within the same chapter. Predicted labels are assigned by the common GPT-5 textual post-processing and taxonomy-mapping stage (Section~\ref{sec:bench-protocol}), which is applied only to model outputs, not to reference annotations; an event mapped to unmapped is excluded from matches. Event F1 (Eq.~\ref{eq:ef1}) is computed over the 16 scored types. Each predicted event interval must lie within its enclosing chapter interval. Although chapter boundaries are shared across systems, predicted event boundaries are generated independently within each chapter (Appendix~\ref{app:metrics}). Thus mIoU, the mean tIoU of matched events, measures the temporal overlap of model-predicted events with matched reference events; it is not determined solely by the common chapterisation. A system matching fewer events can still score a high mIoU, so both are read together.

\paragraph{Unsupported events (T3).} Each predicted event $e\in\hat{\mathcal{E}}$ is labelled as supported or unsupported by the footage, $s(e)\in\{\textsc{sup},\textsc{uns}\}$, by an automated judge validated against human labels (Section~\ref{sec:judge}). The unsupported-event rate is
\begin{equation}
U=\frac{|\{e\in\hat{\mathcal{E}}: s(e)=\textsc{uns}\}|}{|\hat{\mathcal{E}}|}.
\label{eq:unsup}
\end{equation}
The unsupported-event rate is computed over all predicted event records, including multiple events from the same chapter. The chapter identifier is retained for provenance but does not affect the denominator. Unsupported events are the errors most likely to enter a record as fact. Because $U$ can be lowered by reporting fewer events, it is read with event F1; we do not report a risk--coverage analysis.

\paragraph{Audio and speech (T4).} Word error rate (WER) compares each model's transcript with the human-corrected reference after both are normalised with the Whisper English text normaliser \cite{radford2023whisper}, which lower-cases text, removes punctuation and standardises spellings and numbers; non-speech tags (for example [shouting] or [siren]) are removed, names are scored as spelled in the reference, and unintelligible reference segments are excluded.
Speaker accuracy is the share of reference utterances whose role (officer, member of the public or other) is assigned correctly. Each reference utterance is aligned to the predicted utterance with the greatest temporal overlap, and an utterance with no overlapping prediction counts as an error. Where speech overlaps, each reference utterance is scored separately and a predicted utterance can be aligned to only one of them. Reference utterances whose speaker the annotators marked as unknown are excluded.

We do not combine these metrics into a single score. A composite would allow accurate transcription to mask unsupported events, the aggregation a task-grounded approach avoids.

\subsubsection{Automated judge and human validation}
\label{sec:judge}
Support is labelled for every predicted event, after GPT-5 post-processing, by a language-model judge, Doubao Seed 2.0 Pro \cite{bytedance2026seed20}, given the recording's video and audio and one normalised event with its interval, actor and target roles and description (Appendix~\ref{app:prompts}). The judge is a different model from GPT-5, the evaluated systems and the pre-annotation models, but may share failure modes with them; because such judges have documented biases \cite{zheng2023judging,wang2024fair}, we validate it against human labels, and it is excluded from the evaluated panel.
We validated event support using a blinded, stratified human-annotation protocol. The validation sample contained 100 predicted events, 25 from each of the four pipelines run with audio (C2). Two annotators, blind to model identity, reference annotation and judge output, labelled each prediction against the video and audio, with disagreements adjudicated (Appendix~\ref{app:judge}). For the primary binary analysis, supported events were treated as supported; partially supported, uncertain and duplicate events were treated conservatively as unsupported, the same rule the judge applies. We report the confusion matrix, class-wise precision and recall, macro-F1 and Cohen's $\kappa$ \cite{cohen1960kappa}. Support judgements are made independently of reference annotations. The sample is a diagnostic sample, not an estimate of prevalence: its precision and recall are conditional on its stratified composition, agreement between the two annotators before adjudication is not reported, and the rates in Table~\ref{tab:bench-main} come from the judge's labels on all predicted events (Appendix~\ref{app:judge}).

\subsubsection{Analysis plan}
\label{sec:analysis}
With 27 videos we report descriptive comparisons rather than significance tests: the main table averages event F1 over release conditions weighted by video count, and the condition breakdown pools events within each condition. We do not disaggregate by demographic attributes, because inferring them from footage raises known concerns \cite{buolamwini2018gender}.

\section{Experiments and Results}
\label{sec:results}
This section reports the object-level results for Case A (RQ2), the benchmark of event-level pipelines for Case B (RQ3), the validation of the automated judge, and the cross-case comparison.

\subsection{Case A: object-level knife retrieval (E1)}
\label{sec:res-object}
EVA02-Base was the strongest of the four backbones with every protocol (Table~\ref{tab:kh-backbones}, Appendix~\ref{app:kh-ablation}), so Table~\ref{tab:kh-results} compares aggregation heads on it.

\begin{table}[tbp]
\caption{Knife retrieval mAP (\%) on \KH{} with the EVA02-Base backbone. +D: with about 125,000 distractor images.}
\label{tab:kh-results}
\footnotesize
\begin{tabular}{lrrrr}
\toprule
Aggregation & Medium & Hard & Medium+D & Hard+D \\
\midrule
GeMNet \cite{radenovic2019gem} & 83.1 & 61.2 & 81.5 & 57.9 \\
ActNet \cite{husain2021actnet} & 83.9 & 62.1 & 82.6 & 58.9 \\
TokenNet \cite{wu2022tokennet} & 84.8 & 62.5 & 83.5 & 59.8 \\
SENet \cite{lee2023senet} & 85.6 & 63.2 & 84.3 & 61.5 \\
KHNet (ours) & \textbf{86.5} & \textbf{65.3} & \textbf{85.7} & \textbf{63.4} \\
\bottomrule
\end{tabular}
\end{table}

Three observations follow. First, query difficulty dominates model choice: KHNet mAP falls 21.2 points from Medium to Hard (86.5 to 65.3), while its gain over GeMNet on the same backbone is 4.1 points on Hard. Second, model choice matters most in the hardest condition: on Hard+D, KHNet exceeds GeMNet by 5.5 points and SENet by 1.9, and the backbone matters even more, with EVA02-Base 12.4 points above EfficientNet-B7 under GeMNet. Third, distractors cost KHNet 0.8 points on Medium and 1.9 on Hard, so gallery scale matters more when query evidence is weaker. Among the components KHNet adds to the backbone, the ablation (Appendix~\ref{app:kh-ablation}) shows the largest gain on Hard+D from orthogonal fusion of local and global features (58.9 to 62.1). Class activation maps show attention on the blade, the handle or both, and a failure mode in which activation falls on other metallic objects such as axes or knuckle-dusters.

These results have a direct bearing on accountability. The Hard queries come from investigations, which is where a wrong match matters most, and there mAP is 21 to 22 points below the Medium figure that would usually be quoted. The deployment record shows that the system is used, not how often its top match is correct in the field. Open-set queries, calibration of match scores and near-duplicate control between database and queries are not analysed here. A wrong top match would carry wrong metadata, including the offensive-weapon flag, into an intelligence product, so these results support condition-specific reporting and an analyst between the ranked list and any product.

\subsection{Case B: BWV pipeline benchmark}
\label{sec:benchmark}
This benchmark addresses RQ3. It compares complete pipelines, not base models, on all 27 \BWVP{} recordings, without any training on BWV. It uses the event-level metrics of Section~\ref{sec:metrics}, and the protocol was fixed before evaluation.

\subsubsection{Model panel}
\label{sec:bench-panel}
The panel (Table~\ref{tab:bench-panel}, Appendix~\ref{app:bench-tables}) compares our hybrid pipeline with three omni-modal models released in 2026 that accept video with its audio track. They were selected to span the access options open to a police service, from open weights that can run on premises to hosted APIs.
\emph{Our hybrid pipeline}, NarrativeBridge+PAL with GPT-5 post-processing, runs two components in parallel on each recording: NarrativeBridge's cause-effect network \cite{nadeem2025narrativebridge} on the frames, producing timestamped causal-temporal event descriptions within each supplied chapter, and PAL \cite{alex2025pal}, with a Llama 3.2 3B language model, on the audio track, producing a transcription. NarrativeBridge and PAL run locally. Their outputs then pass through the common GPT-5 textual post-processing and taxonomy-mapping stage (Section~\ref{sec:bench-protocol}), which here also combines them. NarrativeBridge alone is the vision-only comparison. \emph{Qwen3.5-Omni-Plus} \cite{qwen2026qwen35omni} supports up to 400 seconds of 720p video at 1 frame per second. \emph{Nemotron 3 Nano Omni} \cite{nvidia2026nemotronomni} is an open-weight 30B-A3B mixture-of-experts model and the only general-purpose base model in the panel that can run on premises. \emph{Gemini 3.1 Pro} \cite{google2026gemini31} is accessed through an API and accepts up to 1M input tokens. Although some models could take longer inputs, every system processes recordings in the same 120\,s windows (Appendix~\ref{app:bench-protocol}), so all receive the same frames.

Appendix~\ref{app:bench-tables} gives the reported results that motivated the choice of NarrativeBridge and PAL.

 The two models used for pre-annotation (Section~\ref{sec:bwv-data}) and the judge model (Section~\ref{sec:judge}) are excluded from the panel.

\subsubsection{Input conditions}
\label{sec:bench-conditions}
Models are evaluated under two input conditions: vision only (C1) and vision with the native audio track (C2). NarrativeBridge covers C1 and the hybrid pipeline covers C2. The comparison measures the effect of adding the PAL audio branch and subsequent fusion within our pipeline; it does not isolate audio, since the PAL transcript also informs GPT-5's mapping and the shared chapters in both conditions come from video and audio.

\subsubsection{Protocol and controls}
\label{sec:bench-protocol}
All systems receive the same fixed chapters (Section~\ref{sec:caseB}), but they independently generate zero or more event claims within each chapter. Each chapter is a shared context interval; each system may produce multiple event claims within it, each with an interval, description and label and role proposals, plus a transcript. The common GPT-5 \cite{openai2025gpt5} textual post-processing and taxonomy-mapping stage applies the same prompt to every base system. It receives these outputs, the transcript, and the 16 event types, but not the original recording. It may revise each event's label and roles, but cannot create, delete, merge or split events or change event or chapter timestamps. Results therefore compare complete base-system-plus-post-processing pipelines rather than raw base models in isolation (prompt in Appendix~\ref{app:prompts}). The GPT-5 stage may normalise outputs towards a shared style, correct or introduce label errors and narrow differences between base systems. Systems also differ in native input handling, context length, windowing, decoding defaults and access mode, and each was run once, so run-to-run variation is not measured. The ARC-Chapter chapters are part of the benchmark condition. We report no ablations without post-processing, without NarrativeBridge or with alternative chapters. Inference settings, windowing and refusal handling are given in Appendix~\ref{app:bench-protocol}.

\paragraph{Reported metrics.} WER, speaker accuracy and mIoU are defined in Section~\ref{sec:metrics}. Event \emph{F1} uses one-to-one matching at $\mathrm{tIoU}\ge0.3$ (Eq.~\ref{eq:ef1}), averaged over release conditions weighted by video count, over the 16 scored event types. \emph{Unsupported events} is $U$ (Eq.~\ref{eq:unsup}). \emph{Latency} is wall-clock time from submitting a recording to receiving the complete post-processed output, per minute of footage, so it includes the shared ARC-Chapter run and the GPT-5 step for every system. Local models run on one workstation (Appendix~\ref{app:bench-protocol});
 for hosted models it includes network transfer and queueing, which we do not control (endpoint regions in Appendix~\ref{app:bench-protocol}). Latency is therefore indicative and not directly comparable between local and hosted models.

\paragraph{Hypotheses.} Stated before evaluation: (H1) Performance falls on edited, long and noisy recordings. (H2) The hybrid pipeline outperforms the general omni models on event detection and speaker attribution.

\subsubsection{Results}
\label{sec:bench-results}
\paragraph{Overall.} On this pilot set, no pipeline is reliable enough to stand in for a reviewer. Even the strongest leaves 13.6\% of predicted events unsupported by the footage and misattributes 27.7\% of utterances, and the general models leave up to 29.5\% of events unsupported (Table~\ref{tab:bench-main}). Differences are descriptive; we report neither significance tests nor confidence intervals. Our hybrid pipeline is best on five of the six measures: the lowest WER (5.7\% against 7.9\% for Gemini 3.1 Pro), the highest speaker accuracy (72.3\% against 69.2\%), the highest event F1 (63.1 against 60.6), the best temporal localisation (mIoU 0.44 against 0.41 for Qwen3.5-Omni-Plus) and the fewest unsupported events (13.6\% against 13.9\% without PAL). Qwen3.5-Omni-Plus records the lowest latency (24.1 against 37.3 seconds per minute of footage), although latency is not directly comparable between local and hosted models. Without the PAL branch (C1), F1 falls to 56.5, below Gemini, and mIoU to 0.39.

\begin{table*}[t]
\caption{Main results on \BWVP{} (27 videos). WER: word error rate against the human-corrected transcript. Speaker: speaker accuracy. F1: event F1 at $\mathrm{tIoU}\ge0.3$, weighted by video count over the release conditions of Table~\ref{tab:bench-conditions} (22 pure BWV, 3 with overlays, 2 with other sources). n/a: not applicable without audio. mIoU: temporal IoU of matched predicted and reference events; chapters are shared context intervals, but event intervals are generated independently within chapters. Unsup.: share of predicted events the validated judge labels unsupported by the footage (Eq.~\ref{eq:unsup}). Latency: wall-clock seconds per minute of footage, including the shared chapterisation run, the GPT-5 post-processing step and network and queueing for hosted API calls, so not directly comparable across access modes. Best per column in bold. Models above the line use condition C2 (vision with audio); the last row is NarrativeBridge with GPT-5 post-processing but without the PAL audio branch (C1). All rows pass through the same GPT-5 post-processing.}
\label{tab:bench-main}
\footnotesize
\begin{tabular}{lcccccc}
\toprule
Model & WER\,\%$\downarrow$ & Speaker\,\%$\uparrow$ & F1$\uparrow$ & mIoU$\uparrow$ & Unsup.\,\%$\downarrow$ & Latency$\downarrow$ \\
\midrule
Hybrid pipeline (ours) & \textbf{5.7} & \textbf{72.3} & \textbf{63.1} & \textbf{0.44} & \textbf{13.6} & 37.3 \\
Gemini 3.1 Pro & 7.9 & 69.2 & 60.6 & 0.35 & 14.2 & 38.5 \\
Qwen3.5-Omni-Plus & 12.5 & 63.8 & 52.0 & 0.41 & 21.7 & \textbf{24.1} \\
Nemotron 3 Nano Omni & 13.7 & 52.4 & 49.7 & 0.38 & 29.5 & 29.8 \\
\midrule
NarrativeBridge + GPT-5 (C1) & n/a & n/a & 56.5 & 0.39 & 13.9 & 37.9 \\
\bottomrule
\end{tabular}
\end{table*}

\paragraph{Results by recording condition.} Table~\ref{tab:bench-conditions} breaks event F1 at $\mathrm{tIoU}\ge0.3$ down by how the recording was released, duration and audio; for edited releases only the retained BWV was scored. The hybrid pipeline scores highest in every condition, descriptively (2 to 22 videos per condition), and all pipelines degrade in the same places: recordings released with overlays score 9.3 (hybrid pipeline) to 12.2 (Qwen) points lower even after the overlaid segments were removed, and long recordings score 12.7 (Gemini) to 16.9 (Qwen) points below short ones. Recordings dominated by shouting or crowd noise are the hardest for every model (37.8 to 49.7).

\begin{table}[tbp]
\caption{Event F1 (\%) at $\mathrm{tIoU}\ge0.3$ on \BWVP{} by recording condition, pooled over events.}
\label{tab:bench-conditions}
\footnotesize
\begin{tabular}{lccccc}
\toprule
Condition & Videos & Hybrid & Gemini & Qwen & Nemotron \\
\midrule
Released as pure BWV & 22 & \textbf{64.7} & 62.4 & 54.0 & 51.6 \\
Released with overlays & 3 & \textbf{55.4} & 51.3 & 41.8 & 39.6 \\
Released with other sources & 2 & \textbf{56.7} & 55.0 & 44.6 & 44.3 \\
\midrule
Short ($<2$\,min) & 8 & \textbf{68.3} & 64.8 & 57.2 & 53.6 \\
Medium (2--5\,min) & 12 & \textbf{58.9} & 50.4 & 42.1 & 40.8 \\
Long ($>5$\,min) & 7 & \textbf{53.0} & 52.1 & 40.3 & 38.4 \\
\midrule
Mostly shouting or crowd audio & 3 & \textbf{49.7} & 44.9 & 38.2 & 37.8 \\
\bottomrule
\end{tabular}
\end{table}

\paragraph{Findings.} On this sample, the results are consistent with H1 for recordings from edited releases, long recordings and noisy audio, and with H2 for event detection and speaker attribution; 27 videos cannot confirm either in general.

\subsection{Judge validation}
\label{sec:res-judge}
Under the conservative rule the 100 validation events comprise 75 supported and 25 unsupported (15 unsupported, 5 partially supported and 5 uncertain), so the uncertainty rate is 5\%. The judge agrees with the settled human labels on 96 events ($\kappa=0.90$, 95\% bootstrap CI 0.78--0.98; Appendix~\ref{app:judge}). For supported events, precision is 98.6\% and recall 96.0\%; for unsupported events, recall is 96.0\% and precision 88.9\%; macro-F1 is 0.95. The false-positive rate is 4.0\% (Wilson 95\% CI 0.7--19.5\%): one of the 25 human-unsupported events was accepted as supported, the error that would let an unsupported statement through, and with 25 such events the judge could plausibly accept up to one in five. The lower unsupported precision reflects class imbalance, since the three supported events the judge rejected count against a class of only 25. The rates in Table~\ref{tab:bench-main} therefore rest on a validated but imperfect instrument, whose error on unsupported events is poorly constrained and was not assessed by event type, condition or model.

\subsection{Cross-case comparison (E3)}
\label{sec:crosscase}
The cases differ in ground truth, metric maturity, field validation, the human role and who is affected by errors (Table~\ref{tab:crosscase}).

\begin{table*}[t]
\caption{Analytical comparison of the two cases.}
\label{tab:crosscase}
\small
\begin{tabular}{p{0.2\linewidth}p{0.35\linewidth}p{0.35\linewidth}}
\toprule
Property & Case A: knife retrieval & Case B: BWV understanding \\
\midrule
Ground truth & Catalogue identity; objective and verifiable & Structured interpretation of an interaction; partly contestable \\
Output form & Ranked list of catalogue items & Free text with intervals \\
Metric maturity & Established retrieval metric (mAP) & Task metrics defined in this paper; support labelled by a judge validated against blinded human labels \\
Field validation & Deployed by UK forces; field accuracy not measured & None; experimental, not used by police; pilot benchmark on public footage only \\
Human role & Analyst confirms top match & Reviewer must verify each event used \\
Main error risk & Plausible wrong match under occlusion or open-set conditions & Fluent, unsupported or misattributed statements \\
Affected parties & Mainly intelligence products about objects & Officers and members of the public described in footage \\
\bottomrule
\end{tabular}
\end{table*}

\section{Discussion}
\label{sec:discussion}

\subsection{Positioning}
Computer vision has documented ties to military and state surveillance \cite{garcia2026weaponization}, and evaluating models for police use risks lending them legitimacy. Here no task identifies individuals and speaker attribution is by role. Because auditing policies can rest on flawed assumptions \cite{escher2026auditing} and benchmark metrics shift \cite{baack2026unsteady}, we evaluate per task and condition, informed by a police partner \cite{ernesto2026proxy}; affected communities and civil-liberties or defence-law experts were not involved, which narrows the perspectives behind the tasks.

\subsection{What the protocol shows about accountability}
\label{sec:implications}
We answer seven questions the measurements raise; the answers are interpretations (E3) and, for the experimental BWV pipeline, describe what would follow if its outputs were adopted.

\paragraph{The institutional failure.} BWV was adopted to make police conduct reviewable, but footage outgrows review. AI tools proposed to close that gap have been judged on acceptance and time saved, not on whether outputs are true to the evidence \cite{white2025truleo,adams2024nomanshand}. The protocol shows what this hides: models that rank well still leave 13.6\% to 29.5\% of predicted events unsupported and misattribute 27.7\% to 47.6\% of utterances, errors that would pass into records unless a workflow required someone to catch them.

\paragraph{Who is burdened, and which records can change.} If adopted, such systems would give institutions scale, reviewers would verify every output, and the people in the footage, who rarely see output or recording, would bear the cost of what is missed. Unsupported outputs could enter incident reports, prosecution case summaries, use-of-force and misconduct records, disclosure schedules and exhibit and knife-intelligence records.

\paragraph{How a claim can be contested.} Contesting an AI-generated claim requires knowing that it was generated, its cited interval and footage, the model version and output log, and the condition's error rates. An analyst can compare a \KH{} match with the physical knife. If BWV outputs were used, an officer could check a transcript against the recording, a prosecutor could request the anchored segment, a defence solicitor could seek disclosure of outputs and error rates, and a member of the public could use the complaints process. The protocol supplies the anchors and per-condition error rates these steps need; whether workflows disclose them is not measured.

\paragraph{Power and procedure.} Automated review would add capacity mainly on one side of an existing evidential asymmetry: the police hold the footage, outputs and logs, while defendants, witnesses and complainants see them only through disclosure. Event labels can lend a contested police account an appearance of objectivity, and the same tools could extend surveillance rather than support review. Accountability would depend on governance we do not evaluate: access to outputs and logs, retention, notice to affected people, correction and appeal, and independent audit of error rates. Data-driven policing can encode existing disparities \cite{richardson2019dirty}, and speech recognition errors differ across speaker groups \cite{koenecke2020asr}; not annotating demographic attributes protects privacy but leaves disparate impact unevaluated, so finding no disparity here is not evidence of none.

\paragraph{What ``support'' means.} Support is relative to the recording, which is partial, recorded from the officer's viewpoint, often trimmed or re-encoded by its publisher (24 of 27 from news outlets) and noisy. A supported event is faithful to the record, not necessarily to what happened.

\paragraph{Does anchoring reduce automation bias?} Anchoring makes checking cheap but can make an incorrect output look checked, as can the automated support check (4.0\% false-positive rate). Reviewer behaviour was not measured, so anchoring is not shown to prevent over-reliance \cite{skitka1999automation,green2019disparate}.

\paragraph{Which errors harm whom.} In records, the worst errors would be, for suspects, an unsupported act of aggression or another speaker's threats attributed to them; for officers, unsupported descriptions of force; for witnesses, misattributed statements; and for communities, degradation on shouting or crowd audio (event F1 37.8 to 49.7), as at protests.

\section{Applications in Policing}
\label{sec:applications}
\KH{} is deployed; the BWV pipeline has not been used or trialled by police, so its uses are prospective. We propose a readiness level for each task, as a governance proposal rather than a validated workflow: \emph{Assistive}, where a person checks the output before relying on it, or \emph{Verify}, where every output used is checked against the source (E3).

\paragraph{Knife identification (T1, Assistive).} An officer photographs a recovered knife and receives ranked catalogue matches, supporting exhibit records, retailer tracing and trend analysis. With 63.4\% mAP on hard queries and field accuracy unmeasured, an analyst confirms each match, a lead, not evidence.

\begin{figure*}[t]
\centering
\resizebox{\textwidth}{!}{%
\begin{tikzpicture}[
  box/.style={draw, rounded corners=2pt, align=center, font=\small, minimum height=1.0cm, text width=2.5cm, fill=white},
  hum/.style={box, fill=gray!15},
  arr/.style={-{Stealth[length=2mm]}, thick},
  lane/.style={font=\small\bfseries, anchor=west},
  note/.style={font=\footnotesize\itshape, align=center}
]
\node[lane] at (-1.4,2.0) {Event level: body-worn video review};
\node[box] (v) at (0,0) {BWV recording\\(frames, audio)};
\node[box] (nb) at (3.2,0.75) {NarrativeBridge\\(frames)\\causal-temporal events};
\node[box] (pal) at (3.2,-0.75) {PAL\\(audio)\\transcription};
\node[box] (gpt) at (6.4,0) {GPT-5\\post-\\processing};
\node[box] (ev) at (9.6,0) {Timestamped events,\\speaker-attributed\\transcript};
\node[hum] (rev) at (12.8,0) {Reviewer verifies\\against footage\\(\emph{Verify})};
\draw[arr] (v.east) -- ++(0.35,0) |- (nb.west);
\draw[arr] (v.east) -- ++(0.35,0) |- (pal.west);
\draw[arr] (nb.east) -- ++(0.35,0) |- (gpt.west);
\draw[arr] (pal.east) -- ++(0.35,0) |- (gpt.west);
\draw[arr] (gpt) -- (ev); \draw[arr] (ev) -- (rev);
\node[note, above=0.05cm of ev] {event F1, mIoU, WER,\\speaker accuracy, $U$};
\node[font=\small\bfseries, align=center, text width=3.2cm] at (0,-4.0) {Object level:\\knife\\identification};
\node[box] (img) at (3.2,-4.0) {Photograph of\\recovered knife};
\node[box] (kh) at (6.4,-4.0) {KHNet\\descriptor};
\node[box] (rank) at (9.6,-4.0) {Ranked catalogue\\matches with\\metadata};
\node[hum] (an) at (12.8,-4.0) {Analyst confirms\\match\\(\emph{Assistive})};
\draw[arr] (img) -- (kh); \draw[arr] (kh) -- (rank); \draw[arr] (rank) -- (an);
\node[note, below=0.05cm of rank] {mAP};
\draw[arr, dashed] (rev.south) -- ++(0,-1.2) -| node[note, pos=0.25, below] {verified event: knife recovered at time $t$} (img.north);
\node[hum, text width=2.6cm] (rec) at (16.1,-2.0) {Linked case record:\\incident account\\and exhibit identity};
\draw[arr] (rev.east) -- ++(0.3,0) |- (rec.west);
\draw[arr] (an.east) -- ++(0.3,0) |- (rec.west);
\end{tikzpicture}
\vspace{-0.3cm}
}
\caption{Illustrative application to a knife-related incident. Event level: NarrativeBridge and PAL run in parallel on the frames and the audio of a BWV recording, and GPT-5 fuses their text outputs, without adding events or changing timestamps, into one record of events and speaker-attributed utterances, which a reviewer verifies against the footage. A verified event in which a knife is recovered prompts a photograph of the knife. Object level: KHNet ranks catalogue matches, which an analyst confirms. Both feed a case record that links the account of the incident to the exhibit. Shaded boxes are human decisions. The BWV pipeline is experimental and has not been used by police, and the combined pipeline has not been evaluated.}
\Description{Two horizontal lanes. The upper lane, event level, starts with a body-worn video recording that splits into NarrativeBridge on the frames and PAL on the audio; their text outputs are fused by GPT-5, without adding events or changing timestamps, into timestamped events and a speaker-attributed transcript, evaluated by event F1, mIoU, word error rate, speaker accuracy and unsupported-event rate, and passed to a reviewer who verifies them against the footage. A dashed arrow from the reviewer, labelled verified event: knife recovered at time t, leads down to the lower lane, object level, which runs from a photograph of the recovered knife to the KHNet descriptor, to ranked catalogue matches with metadata evaluated by mAP, to an analyst who confirms the match. Arrows from the reviewer and the analyst join at a linked case record holding the incident account and the exhibit identity.}
\label{fig:application}
\vspace{-0.5cm}
\end{figure*}
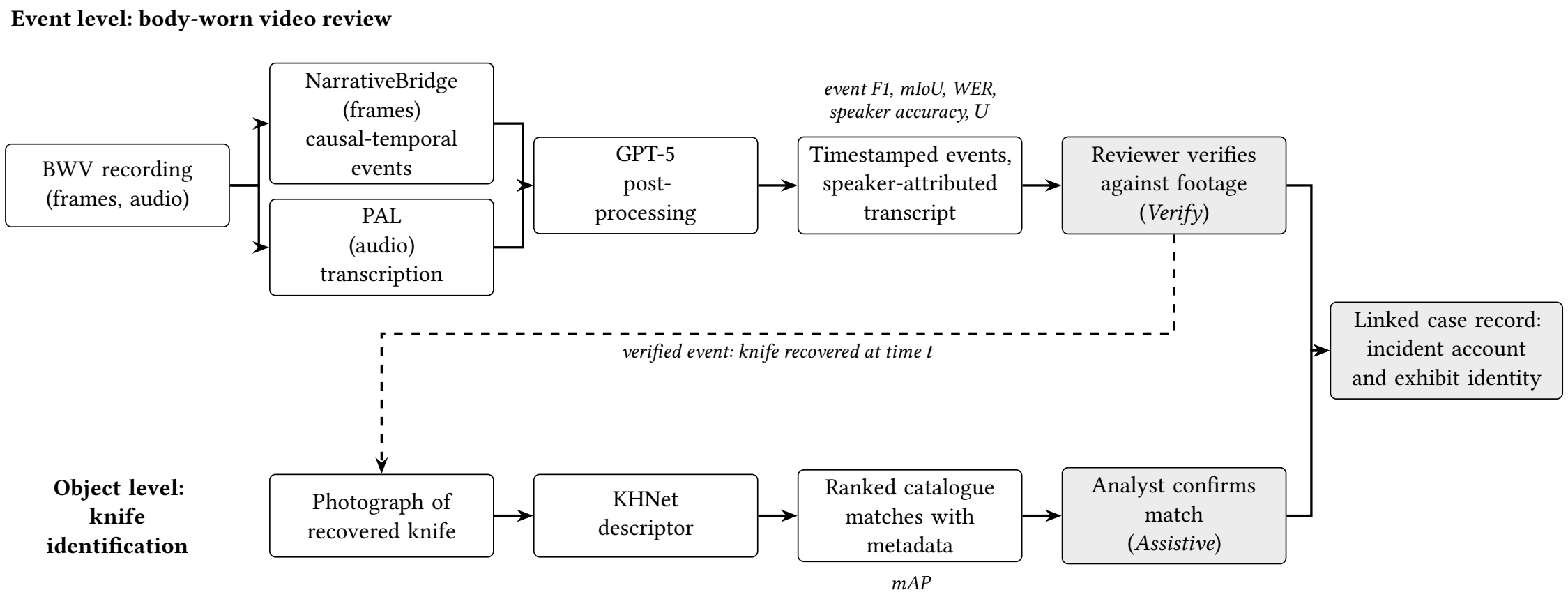

\paragraph{BWV review (T2--T4, Verify).} Event-level outputs could reduce, not replace, watching footage. Timestamped events would let reviewers jump to a search, arrest or use of force, transcripts would start case files, and footage could be sampled for supervision, not only after complaints. Given these error rates, every output entering a record would need verifying against the footage, which may shift rather than remove the review burden; verification time, workload and error detection were not measured. Outputs should not inform probable cause, arrest, force review, disciplinary findings or evidential conclusions without independent human and legal review. Figure~\ref{fig:application} illustrates how event detection could locate a knife for \KH{} to identify; this combined use is unevaluated.

\section{Limitations}
\label{sec:limitations}
(1) \BWVP{} is a small pilot set of selected, publicly released footage; it supports descriptive comparison, not generalisation, also informed prompt design, and we report no confidence intervals or per-recording results for the benchmark. (2) References are machine-assisted and human-corrected, without independent annotation or agreement statistics; WER may favour transcripts resembling Whisper's \cite{mei2026aphasia}. (3) The judge was validated on 100 stratified events, not at natural prevalence or by event type and condition. (4) Pipelines were run once, without ablations of post-processing, components or chapters, and errors are not attributed to stages. (5) Contamination by public videos and reports is possible. (6) There is no reviewer study, affected-party input, per-event-type analysis or cost and energy accounting. (7) \KH{}'s field accuracy, open-set queries and calibration are not evaluated. (8) Group differences are unmeasured.

\vspace{-0.3cm}
\section{Conclusion and Future Work}
\label{sec:conclusion}
We argued that advances in video AI could make CV a realistic aid to BWV review if its claims are traceable, its failures visible and its outputs contestable, and contributed an evidence-grounded protocol that measures output-level evidence for these conditions at two levels. At the object level, KHNet reaches 85.7\% and 63.4\% mAP on the medium and hard \KH{} protocols with distractors, a 22.3-point drop. On the \BWVP{} pilot set, our hybrid pipeline leads on five of six measures (event F1 63.1 against 60.6 for Gemini 3.1 Pro; 56.5 without PAL), suggesting that such analysis could be useful for policing. Yet even the best pipeline leaves 13.6\% of events unsupported by the recording, which is not the same as untrue, and misattributes 27.7\% of utterances. Future work: independent annotation with agreement statistics, a larger systematically sampled set, judge validation at natural prevalence, pipeline ablations, a reviewer study and consultation with affected communities.

\vspace{-0.3cm}
\begin{acks}
This work was supported by the Open-Source AI Fellowship, a partnership between The Alan Turing Institute, the Incubator for AI (i.AI) and Meta.
\end{acks}

\section*{Ethical Considerations Statement}
This work concerns AI for police work, which affects officers and members of the public who have not consented to being analysed. The study was reviewed by the Incubator for AI (i.AI) and the University of Surrey, neither of which reported ethical concerns.

\emph{Data.} We collected no new data from people and used no police-held footage: every BWV recording was already published by a police force or news outlet. Public availability is not consent, especially for victims, young people and people filmed at traumatic moments, so we name no one, infer no demographic attributes and redistribute no video, releasing only source links and annotations. A recording withdrawn by its publisher is removed at the next release, with results recomputed, and people shown can request removal or correction of annotations (Appendix~\ref{app:datasheet}). The knife dataset contains images of objects, not people.

\emph{Hosted models.} Only public footage was sent to providers, through the regions listed in Appendix~\ref{app:bench-protocol}, under each provider's standard terms. Gemini 3.1 Pro and GPT-5 were used through paid accounts, whose terms exclude training on API data \cite{google2026geminiterms,openai2026apidata}; GPT-5 received text only. Qwen3.5-Omni-Plus and the judge were used mainly through free quotas; Alibaba Cloud states that it retains inference data for at most 30 days and does not train on it without consent \cite{alibaba2026retention}, and we did not verify whether free use changes these terms.

\emph{Risks.} Benchmark results could be used to justify automating decisions that should stay human, or to extend surveillance beyond accountability. We therefore report results per task, keep all event-level outputs at the \emph{Verify} level (Section~\ref{sec:applications}), and state that outputs should not inform arrests, force reviews, disciplinary findings or evidential conclusions without independent human and legal review. Because system errors could change the account of what people in the footage did, any deployment would need notice, access and correction mechanisms for them.
\vspace{-0.3cm}
\section*{Generative AI Usage Statement}
A generative AI assistant was used to help structure and draft sections of this manuscript and to format the bibliography. All technical content, results and citations were checked by the authors against primary sources, and the authors take full responsibility for the paper.

\bibliographystyle{ACM-Reference-Format}
\bibliography{references}

\appendix

\section{Annotation Guidelines}
\label{app:guidelines}
\paragraph{Event types.} Each event is labelled with one of the 16 scored types in Table~\ref{tab:inventory}. Every base system's predicted events are mapped to one of the 16 scored types, or to unmapped, by the common GPT-5 textual post-processing and taxonomy-mapping stage (Section~\ref{sec:bench-protocol}), which is not given type~9, so a predicted and a reference event have equivalent labels when their types are the same. Type~9 is a video-level tag, not an event type: it is recorded from the publication context in the datasheet, is not annotated as an interval, and is excluded from event matching, event F1 and the unsupported-event rate.

\begin{table*}[h]
\caption{Event types in \BWVP{}, with the recordings in which each occurs. Type~9 (shaded) is a video-level tag taken from the publication context and is not scored.}
\label{tab:inventory}
\footnotesize
\begin{tabular}{p{0.03\linewidth}p{0.24\linewidth}p{0.42\linewidth}p{0.22\linewidth}}
\toprule
\# & Event type & What it covers & Videos \\
\midrule
1 & Armed suspect confrontation (edged or blunt weapon) & Officers face someone holding a knife, hammer or similar, with shouted commands and disarming & uk01, uk04, uk08, uk12, uk14, uk15, uk22, uk25 \\
2 & Firearms incident & Armed police stops and discharges, or arrests of suspects linked to firearms & uk16, uk23 \\
3 & Taser use & Taser drawn or fired to stop a suspect & uk01, uk14, uk24, uk25 \\
4 & Assault on officers & An officer is attacked during the encounter & uk07, uk08, uk12, uk15 \\
5 & Arrest at a live scene & Arrest straight after a violent incident, while the scene is still unfolding & uk01, uk04, uk10, uk14, uk22, uk25 \\
6 & Planned arrest at a home & Officers arrive at an address, read the caution and detain the person & uk17, uk19, uk20 \\
7 & Arrest after a search or manhunt & Suspect found and arrested in public after a search & uk18, uk23 \\
8 & Handcuffing and physical restraint & Cuffing, holding on the ground, prolonged restraint & uk03, uk04, uk05, uk13, uk21, uk26, uk27 \\
\rowcolor{gray!15} 9 & Contested or excessive use of force (video-level tag, not scored) & Force later challenged at court, a misconduct hearing or an inquest & uk03, uk07, uk21, uk24, uk26, uk27 \\
10 & Vehicle pursuit and tactical contact & Chases, ramming to stop a vehicle, boxing in, failure to stop & uk02, uk11, uk15 \\
11 & Foot chase and decamp & Suspect leaves the vehicle and runs; dog or air support tracks them & uk02, uk11 \\
12 & Roadside vehicle stop and search & Car pulled over, occupants removed, searched or questioned & uk16, uk21, uk24 \\
13 & Public disorder & Riot, a fight outside a nightclub, crowd violence & uk09, uk10 \\
14 & Scene interaction with suspects or witnesses & Officers talking to people at a scene, sometimes not yet knowing who the offender is & uk06, uk09, uk22 \\
15 & Injured person or medical emergency & Victim care, first aid, a person in distress while restrained & uk05, uk13, uk27 \\
16 & Victim mistaken for a suspect & The person being restrained turns out to be the victim & uk05, uk13 \\
17 & In-vehicle custody exchange & Suspect in a police car giving details or asking about the cuffs & uk09 \\
\bottomrule
\end{tabular}
\end{table*}

\paragraph{Inclusion and exclusion.} An event is annotated if it is visible in body-worn footage or clearly audible on its soundtrack. Excluded are events known only from narration, captions or on-screen text, and events shown only in news-studio, CCTV or other non-body-worn inserts, which were removed before evaluation. Labels describe what happens, not motive, emotion or lawfulness. Where an edited release repeats the same moment, each occurrence is annotated and the repeat is flagged.

\paragraph{Duration and boundaries.} An event lasts at least 1\,s; an instantaneous action, such as a Taser discharge, is given a 1\,s interval starting at the action. Boundaries are marked to the nearest second, from the first to the last frame or sound in which the event is perceptible. Two events of the same type less than 2\,s apart are merged.

\paragraph{Overlapping events.} Events often overlap or nest, because the types describe different aspects of an encounter. In uk01, for example, an armed confrontation (type~1), Taser use (type~3) and an arrest at a live scene (type~5) occur in sequence and partly overlap, and a restraint (type~8) may involve an injured person (type~15). Each event is annotated with its own interval and matched separately under one-to-one matching.

\paragraph{Borderline cases.} (1) \emph{Arrest types}: an arrest is labelled type~5 when it follows a violent incident still unfolding on camera, type~6 when officers arrive at an address to detain a named person, and type~7 when the person is found after a search shown or referred to in the footage; when two apply, the annotator chooses the type the footage shows most directly. (2) \emph{Restraint versus guiding}: brief contact to guide a compliant person is not labelled; it becomes type~8 only when the officer holds the person so as to restrict movement for at least 1\,s. (3) \emph{Victim mistaken for a suspect}: type~16 is labelled only from the point at which the footage itself shows or states that the restrained person is the victim, for example when an officer says so; before that point the restraint is labelled type~8 only.

\paragraph{Event support.} For each predicted event in the judge validation sample, annotators record its support status (supported, partially supported, unsupported or uncertain), the event type observed (Table~\ref{tab:inventory}), whether its timing is adequate, whether its actor and target roles are correct and whether it contains unsupported detail, without knowing which model produced it or the judge's label.

\paragraph{Source segments.} Before evaluation, segments are marked as clean body-worn footage, body-worn footage with broadcaster overlays, or non-body-worn footage, using only the recording's visual source and editing properties, never the incident type, events, transcript or publication headline. Only clean body-worn segments are retained.

\paragraph{Speaker roles.} Each utterance is assigned to officer, member of the public, other, or unknown. Annotators record ``unknown'' when the speaker cannot be determined from the footage.

\paragraph{Prohibited labels.} Annotators do not record perceived race, ethnicity, gender, age, religion or disability.

\section{Event-Level Metrics}
\label{app:metrics}
\paragraph{What each metric measures.} We keep four properties separate: event F1 combines \emph{detection} and \emph{classification} at a lenient localisation threshold, mIoU measures \emph{temporal localisation}, and $U$ measures \emph{evidential support}, whether a predicted event occurs in the footage at all.
\paragraph{Chapters and events.} Chapter intervals are fixed context boundaries; event intervals are model outputs. The chapteriser does not determine the event interval, so temporal localisation is model-specific because event intervals can differ between systems. Predicted events are pooled across all chapters of a recording before matching, and both predicted and reference events may overlap or nest within a chapter.
\paragraph{Event detection (T2).} A predicted event $\hat e=(\hat\ell,\hat u)$ matches a reference event $e^\ast=(\ell^\ast,u^\ast)$ if $\mathrm{tIoU}(\hat u,u^\ast)\ge\theta$ and $\hat\ell$ and $\ell^\ast$ map to the same event type in Table~\ref{tab:inventory}; the video-level tag (type~9) is excluded. Under one-to-one Hungarian matching,
\begin{equation}
\mathrm{F1}_E=\frac{2P_ER_E}{P_E+R_E},\qquad \theta=0.3.
\label{eq:ef1}
\end{equation}

\section{Benchmark Protocol Details}
\label{app:bench-protocol}
\paragraph{Common chapterisation.} ARC-Chapter \cite{pu2025arcchapter} was run once per video on its frames and audio track and frozen before any comparison system was evaluated. Chapterisation was not repeated per system and received no outputs from evaluated systems. ARC-Chapter was not fine-tuned or prompted with \BWVP{} annotations; it was trained on web videos with creator-provided chapters, so overlap with these public recordings cannot be ruled out, and its boundaries may suit edited, broadcast-style structure better than unedited BWV. Because every predicted event must lie within its enclosing chapter, a reference event that crosses a chapter boundary can be matched only by a predicted event inside one chapter, so chapterisation can limit attainable tIoU equally for every system.

\paragraph{Frames and context.} Frames are sampled uniformly at 1 frame per second with a maximum of $F_{\max}=128$ frames per model request. Recordings longer than 120 seconds are divided into windows of $w=120$ seconds with an overlap of $o=30$ seconds. Thus, each window contains at most 120 sampled frames, leaving eight frame slots below the 128-frame cap; timestamp text and generated output consume separate context-token budget. Window outputs are merged by chapter identifier using global timestamps; chapters cannot be duplicated or re-bounded by windowing, and every event interval is kept within its enclosing chapter. For transcripts, overlapping text is deduplicated using timestamp overlap rather than simple string concatenation. For models without native temporal encoding, each frame is preceded by its global timestamp so that outputs can cite intervals.
\paragraph{Prompting and decoding.} The omni models receive the same instruction and JSON output schema (Appendix~\ref{app:prompts}), with the default decoding settings for each API. NarrativeBridge, alone and within the hybrid pipeline, uses beam search with beam size 5, as in its original paper; GPT-5 post-processing uses the prompt in Appendix~\ref{app:prompts}.
\paragraph{Invalid outputs and refusals.} An output that does not parse against the schema is retried at most twice and then scored as empty. A refusal to describe the footage is not retried with a modified prompt, because prompt changes that work around a model's safety behaviour would not reflect how the model behaves in deployment.
\paragraph{Hardware and endpoints.} Local components ran on one workstation, with its system clock synchronised to UTC so that latencies are timed consistently. In earlier tests on a node with two 24\,GB GPUs, we measured GPU memory use of about 12\,GB for NarrativeBridge's released checkpoint, about 24\,GB with its feature encoders, and at least 20\,GB for PAL with Llama 3.2 3B, so the local part of the hybrid pipeline fits on two 24\,GB GPUs; a fully on-premises version would need a local model in place of the GPT-5 post-processing step, which every system in this benchmark uses. NarrativeBridge, PAL with Llama 3.2 3B and Nemotron 3 Nano Omni (open weights) ran on the local workstation; GPT-5 post-processing used the OpenAI API in the nearest supported region; Gemini 3.1 Pro used the Google API with UK/EU data processing; Qwen3.5-Omni-Plus used the Alibaba Cloud API in Singapore, as no UK or EU region was offered; and the judge, Doubao Seed 2.0 Pro, used its API in the nearest supported region. All judge calls used the same region and were made at varying times rather than in a single run window.

\paragraph{Data handling.} Only publicly released footage is sent to hosted APIs. This is a property of the benchmark, not a recommendation: operational footage would require on-premises models or an approved processing agreement.

\section{Judge Validation Protocol and Sample}
\label{app:judge}
\paragraph{Sample and annotation.} Sampling was stratified by model, recording and event condition and included supported, unsupported, borderline and duplicate predictions. Two annotators independently reviewed the original video and audio, together with the model-predicted event and its temporal interval, without access to the model identity, reference annotation or automated-judge output. They labelled each prediction as supported, partially supported, unsupported or uncertain and recorded the event type (Table~\ref{tab:inventory}), temporal adequacy, whether the actor and target roles were correct and whether the claim contained unsupported detail. Disagreements were adjudicated by discussion and, where necessary, by a third annotator.
\paragraph{Support labels.} \emph{Supported}: the event is directly observable or clearly audible in the footage. \emph{Partially supported}: the general event occurs, but the predicted label includes an unsupported detail or the timing is materially wrong. \emph{Unsupported}: the event or its essential claim is not supported by the footage. \emph{Uncertain}: the footage is too ambiguous, occluded, edited or noisy to decide. Two flags are recorded in addition to the label and are not classes: \emph{borderline}, for cases the annotators found hard to decide, and \emph{duplicate}, for predictions that repeat or overlap another predicted event. Borderline events keep their support label; duplicates are counted as unsupported, the same rule the judge applies.
\paragraph{Resolving disagreements.} After independent annotation, (1) the two labels are compared; (2) disagreements are identified; (3) the annotators discuss only the disagreements; (4) if they reach agreement, the settled label is recorded; (5) if not, a third, senior annotator adjudicates; and (6) the original independent labels are preserved with the settled labels.
\paragraph{Mapping for the binary comparison.} Supported $\rightarrow$ supported; partially supported $\rightarrow$ unsupported; uncertain $\rightarrow$ unsupported, with the uncertainty rate reported separately; duplicate $\rightarrow$ unsupported.
\paragraph{Confusion matrix.} Against the settled human labels, the judge labelled 72 of the 75 supported events supported and 3 unsupported, and 24 of the 25 unsupported events unsupported and 1 supported (73 judged supported, 27 unsupported).

\paragraph{Sample.} Table~\ref{tab:judge-sample} gives the composition of the validation sample. It was stratified by model, recording and event condition and is not intended to estimate the natural prevalence of unsupported events, which is estimated from the judge's labels on all predicted events.

\begin{table}[h]
\caption{Composition of the judge validation sample by settled human label. Borderline and duplicate are flags, not classes, and are not included in the total.}
\label{tab:judge-sample}
\footnotesize
\begin{tabular}{lcccccccc}
\toprule
Model & Videos & Sup. & Unsup. & Partial & Uncertain & Borderline & Duplicate & Total \\
\midrule
Hybrid pipeline & 20 & 19 & 3 & 2 & 1 & 2 & 1 & 25 \\
Gemini 3.1 Pro & 21 & 19 & 4 & 1 & 1 & 2 & 1 & 25 \\
Qwen3.5-Omni-Plus & 19 & 18 & 4 & 1 & 2 & 2 & 0 & 25 \\
Nemotron 3 Nano Omni & 20 & 19 & 4 & 1 & 1 & 2 & 1 & 25 \\
\midrule
Total & -- & 75 & 15 & 5 & 5 & 8 & 3 & 100 \\
\bottomrule
\end{tabular}
\end{table}

\section{Model Selection and Panel}
\label{app:bench-tables}

\paragraph{Why NarrativeBridge and PAL} Each was the strongest option in its own evaluation. NarrativeBridge outperformed task-specific captioners and zero-shot and fine-tuned video-language models on every causal-temporal narrative metric, with CIDEr margins of 17.88 (MSVD-CTN) and 17.44 (MSRVTT-CTN) over the strongest baseline (Table~\ref{tab:nb-reported}) \cite{nadeem2025narrativebridge}. PAL gave the best result on every audio task except VocalSound for all three language-model backbones (Table~\ref{tab:pal-reported}), and its attention-based integration uses about 60\% less memory than prepending audio tokens \cite{alex2025pal}, which matters because footage that cannot leave a police network must be processed on local hardware.

\begin{table}[h]
\caption{Results reported by NarrativeBridge \cite{nadeem2025narrativebridge}: NarrativeBridge against task-specific methods and video-language models on MSVD-CTN and MSRVTT-CTN. R-L, C and S denote ROUGE-L, CIDEr and SPICE.}
\label{tab:nb-reported}
\footnotesize
\begin{tabular}{lcccccc}
\toprule
 & \multicolumn{3}{c}{MSVD-CTN} & \multicolumn{3}{c}{MSRVTT-CTN} \\
\cmidrule(lr){2-4}\cmidrule(lr){5-7}
Method & R-L$\uparrow$ & C$\uparrow$ & S$\uparrow$ & R-L$\uparrow$ & C$\uparrow$ & S$\uparrow$ \\
\midrule
SEM-POS \cite{nadeem2023sempos} & 25.39 & 37.16 & 14.46 & 20.11 & 26.01 & 12.09 \\
AKGNN \cite{hendria2023akgnn} & 25.11 & 35.08 & 14.55 & 21.42 & 25.90 & 11.99 \\
GIT \cite{wang2022git} & 27.51 & 45.63 & 15.58 & 24.51 & 32.43 & 13.70 \\
VideoLLaVA (zero-shot) \cite{lin2024videollava} & 21.80 & 30.55 & 14.67 & 19.33 & 16.24 & 12.49 \\
VideoLLaVA (LoRA FT) & 24.56 & 34.98 & 15.41 & 21.21 & 18.97 & 13.28 \\
VideoLLaVA (simple FT) & 25.61 & 36.12 & 16.09 & 22.18 & 19.98 & 13.07 \\
ShareGPT4Video (zero-shot) \cite{chen2024sharegpt4video} & 21.66 & 27.06 & 14.06 & 20.27 & 17.08 & 12.21 \\
ShareGPT4Video (LoRA FT) & 24.39 & 30.72 & 14.83 & 22.09 & 19.83 & 13.02 \\
ShareGPT4Video (simple FT) & 25.32 & 31.67 & 14.92 & 23.01 & 20.76 & 13.28 \\
\midrule
NarrativeBridge & \textbf{31.46} & \textbf{63.51} & \textbf{19.25} & \textbf{27.90} & \textbf{49.87} & \textbf{15.76} \\
\bottomrule
\end{tabular}
\end{table}

\begin{table*}[h]
\caption{Results reported by PAL \cite{alex2025pal} for three language-model backbones: the standard approach of prepending audio tokens to the language-model input (PLITS) against attention-based integration (LAL), alone and with fused audio tokens, which is PAL (LAL+LFST). Classification: accuracy (ESC-50, VocalSound), micro-F1 (DCASE), mAP (FSD, AudioSet-2M). Captioning on AudioCaps (AC) and Clotho (CL): $\dagger$ CIDEr, $\ddagger$ SPICE. Best per backbone in bold.}
\label{tab:pal-reported}
\footnotesize
\begin{tabular}{llccccccccc}
\toprule
 & & \multicolumn{5}{c}{Classification} & \multicolumn{4}{c}{Captioning} \\
\cmidrule(lr){3-7}\cmidrule(lr){8-11}
Backbone & Integration & ESC50 & DCASE & VS & FSD & AS2M & AC$\dagger$ & CL$\dagger$ & AC$\ddagger$ & CL$\ddagger$ \\
\midrule
Llama3.2-1B & PLITS & 64.45 & 37.69 & 51.57 & 25.23 & 9.08 & 0.59 & 0.34 & 16.30 & 10.96 \\
 & LAL & 76.70 & 40.97 & \textbf{60.87} & 31.44 & 11.83 & 0.66 & 0.38 & 16.97 & 11.87 \\
 & PAL (LAL+LFST) & \textbf{87.40} & \textbf{46.23} & 56.03 & \textbf{43.91} & \textbf{14.74} & \textbf{0.72} & \textbf{0.42} & \textbf{18.08} & \textbf{12.58} \\
\midrule
Llama3.2-3B & PLITS & 70.40 & 40.62 & 61.40 & 28.88 & 10.84 & 0.63 & 0.35 & 16.81 & 11.35 \\
 & LAL & 82.15 & 43.21 & \textbf{65.78} & 34.29 & 12.91 & 0.67 & 0.38 & 17.80 & 12.18 \\
 & PAL (LAL+LFST) & \textbf{89.25} & \textbf{47.21} & 60.46 & \textbf{43.86} & \textbf{15.03} & \textbf{0.73} & \textbf{0.40} & \textbf{18.61} & \textbf{12.46} \\
\midrule
Qwen2.5-1.5B & PLITS & 68.00 & 37.57 & 56.45 & 27.87 & 9.56 & 0.63 & 0.38 & 16.63 & 11.74 \\
 & LAL & 70.85 & 38.79 & \textbf{59.20} & 28.53 & 10.28 & 0.63 & 0.38 & 16.65 & 11.44 \\
 & PAL (LAL+LFST) & \textbf{87.80} & \textbf{45.52} & 56.73 & \textbf{43.26} & \textbf{13.92} & \textbf{0.73} & \textbf{0.41} & \textbf{18.45} & \textbf{12.20} \\
\bottomrule
\end{tabular}
\end{table*}

\begin{table}[h]
\caption{Model panel. Input: V = frames, A = native audio track. n/d: not disclosed by the provider. The NarrativeBridge (CEN) size is estimated from its released 32-bit checkpoint (317\,MB) and excludes the two CLIP4Clip video encoders that supply its features. The hybrid pipeline's NarrativeBridge and PAL components run locally. Every system's outputs pass through the common GPT-5 textual post-processing and taxonomy-mapping stage through an API.}
\label{tab:bench-panel}
\footnotesize
\begin{tabular}{p{0.30\linewidth}p{0.12\linewidth}p{0.17\linewidth}p{0.16\linewidth}p{0.11\linewidth}}
\toprule
Model & Size & Access & Input & Reference \\
\midrule
NarrativeBridge & $\approx$80M & Local & V & \cite{nadeem2025narrativebridge} \\
PAL + Llama 3.2 3B & $\approx$3B & Local & A & \cite{alex2025pal} \\
GPT-5 post-processing (all systems) & n/d & API & Base-system text & \cite{openai2025gpt5} \\
Hybrid pipeline (ours) & -- & Hybrid local/API & V+A & This paper \\
\midrule
Qwen3.5-Omni-Plus & n/d & API & V+A & \cite{qwen2026qwen35omni} \\
Nemotron 3 Nano Omni & 30B-A3B & Open weights & V+A & \cite{nvidia2026nemotronomni} \\
Gemini 3.1 Pro & n/d & API & V+A & \cite{google2026gemini31} \\
\bottomrule
\end{tabular}
\end{table}

\section{Prompt Templates}
\label{app:prompts}
All base systems are passed through the same common GPT-5 textual post-processing and taxonomy-mapping stage before scoring. The reported results therefore measure complete base-system-plus-GPT-5 pipelines. They should not be interpreted as raw performance of the base models in isolation. GPT-5 may interact differently with outputs from different base systems, despite the common prompt and schema. The base systems may use different input prompts internally; the common post-processing prompt, taxonomy, output schema, and evaluation procedure are fixed across systems. Each base system receives the same fixed chapter intervals from the common chapterisation (Section~\ref{sec:caseB}). The prompts below are used only by the omni-modal base systems: using the event-extraction prompt, each outputs zero or more event claims within each chapter, each with an event identifier, an interval inside the chapter, a description, an event-label proposal and an actor/target proposal, together with a transcript, keeping chapter identifiers and timestamps unchanged. The hybrid pipeline takes no prompt at this stage: NarrativeBridge outputs timestamped event descriptions within each chapter and PAL a transcript, and event labels and actor and target roles are assigned only by GPT-5. The post-processing prompt supplies the 16 scored event types of Table~\ref{tab:inventory}, excluding type~9. Some prompts below also contain source-evidence and footage-type fields; because non-BWV and overlaid segments were removed before evaluation, these fields are not used or reported, and the footage-type section of the GPT-5 prompt is omitted where marked. Uncertainty fields from the speech prompt are discarded by the common output schema. The omni-modal base systems were also given the 16 scored types for reference; the labels scored are those assigned by GPT-5. The judge receives the recording and one predicted event, but not the reference annotation, the model identity or any benchmark score. Human annotators in the judge validation use the coding form at the end of this appendix and do not see the judge's output. Model versions: Gemini 3.1 Pro as \texttt{gemini-3.1-pro-preview}; Qwen3.5-Omni-Plus as \texttt{qwen3.5-omni-plus}, equivalent to the snapshot \texttt{qwen3.5-omni-plus-2026-03-15}; Nemotron 3 Nano Omni from its released open weights; GPT-5, used for the common textual post-processing and taxonomy-mapping stage, as \texttt{gpt-5-2025-08-07}; and the judge as Doubao Seed 2.0 Pro. Preview and alias endpoints can change the model they serve, so results reflect the versions served at run time.

\begin{promptbox}{System prompt (omni-modal base systems)}
\begin{verbatim}
You are an evidence-grounded assistant for reviewing body-worn-video recordings.

Your task is to describe observable events, transcribe speech, assign speaker roles, and
classify footage segments. You are not a police decision-maker, legal assessor, or
investigator.

Use only information visible or audible in the supplied recording. Do not infer intent,
guilt, legality, identity, race, ethnicity, age, gender, disability, or mental state. Do not
identify people by name.

If the footage is unclear, edited, obstructed, off-screen, or inaudible, report uncertainty
rather than guessing. Distinguish clearly between:
- directly observed or audible information;
- information that is partially supported;
- information that is not supported;
- information that cannot be determined.

Every event must include a start and end time. Keep event descriptions concise and
evidence-grounded. Do not merge separate events unless they are clearly continuous. Do not
invent events to make the narrative complete.

Return only valid JSON conforming to the supplied schema.
\end{verbatim}
\end{promptbox}

\begin{promptbox}{Event-extraction prompt and output schema (omni-modal base systems)}
\begin{verbatim}
The recording has already been partitioned into fixed intelligent chapters.
Do not create, delete, merge, split, extend, shorten, or retimestamp chapters.

For each supplied chapter, identify zero or more observable events relevant to
evidential review. A chapter may contain multiple events, including events
that overlap or occur in sequence.

For each predicted event:

1. Assign a unique event_id within the recording.
2. Give its start and end time.
3. Keep the event interval within the enclosing chapter interval.
4. Describe one observable action or interaction.
5. Propose one label from the supplied taxonomy, or "unmapped".
6. Identify actor and target only by role where supported:
   officer, member_of_public, other, or unknown.
7. Do not merge distinct events merely because they occur in the same chapter.
8. Do not create an event solely from a transcript utterance.
9. Do not infer intent, legality, guilt, identity, or causation.

Return one record for every predicted event and preserve the enclosing
chapter_id for each event.
\end{verbatim}
\tcbline
\begin{verbatim}
{
  "chapters": [
    {
      "chapter_id": "chapter_001",
      "start_time": 0.0,
      "end_time": 30.0,
      "events": [
        {
          "event_id": "chapter_001_event_001",
          "start_time": 4.0,
          "end_time": 8.0,
          "description": "An officer speaks to a member of the public.",
          "event_label": "scene_interaction_with_suspects_or_witnesses",
          "actor_role": "officer",
          "target_role": "member_of_public"
        },
        {
          "event_id": "chapter_001_event_002",
          "start_time": 12.0,
          "end_time": 18.5,
          "description": "An officer handcuffs the person's wrists behind their back.",
          "event_label": "handcuffing_and_physical_restraint",
          "actor_role": "officer",
          "target_role": "member_of_public"
        }
      ],
      "source_evidence": "Continuous body-worn-camera footage is visible."
    }
  ]
}
\end{verbatim}
\end{promptbox}

\begin{promptbox}{GPT-5 post-processing prompt and common output schema (all systems; abridged)}
\begin{verbatim}
You are a schema-constrained post-processing component in an
evidence-grounded body-worn-video evaluation pipeline.

You receive:

1. records produced by a base video-understanding system for fixed,
   pre-supplied chapters. Each chapter record has:
   - a chapter ID and start and end timestamps;
   - zero or more event proposals, each with an event ID, start and end
     timestamps inside the chapter, a description, a proposed event label,
     and proposed actor and target roles;
   - [optional source or editing evidence, not used in this study].

2. a timestamped transcript produced by the same base system;

3. the 16 scored event types and their definitions;

4. [footage-type labels, not used in this study].

You do not receive the original video or audio. Use only the supplied chapter
outputs, transcript and event taxonomy. Do not use
outside knowledge.

Your task is to normalize the existing chapters, events and transcript into
the common evaluation schema.

CHAPTER PRESERVATION

For every input chapter, preserve exactly one chapter record and its
identifier and timestamps. Within that chapter, preserve every base-system
event proposal and allow multiple event records.

GPT-5 may revise each event's taxonomy label, actor role, and target role
using the description and transcript, but it must not create, delete, merge,
or split events. It must preserve every event_id and every event timestamp.
Every event interval must remain within its enclosing chapter interval.

A chapter may contain zero, one, or multiple events. Do not collapse multiple
events into one chapter-level event. Preserve chapter and event order and
do not alter chapter intervals.

EVENT-TAXONOMY MAPPING

- Map each existing event to exactly one of the 16 scored event types, or to
  "unmapped" if no type's definition is supported by the event.
- Select the type whose definition is best supported by the event
  description and any relevant transcript text. Treat the base system's
  proposed label and roles as inputs to check, not as final.
- Do not create an event merely because it appears in the transcript.
- Do not use outside knowledge, publication context, incident headlines, or
  police-force identity.
- Do not infer legality, guilt, intent, causation, credibility, or identity.
- The taxonomy labels are operational benchmark categories, not legal findings.
- Do not force a label: if no type is supported, use "unmapped" without adding
  unsupported details.

ACTOR AND TARGET ROLES

- Infer actor_role and target_role only from the event description and
  transcript when the relationship is sufficiently supported.
- Allowed values are:
  officer, member_of_public, other, unknown.
- Use unknown when the role cannot be established.
- A speaker role does not automatically establish that the speaker is the
  actor in the event.
- Do not infer personal identity or demographic attributes.

[FOOTAGE-TYPE MAPPING section omitted: not used in this study]

TRANSCRIPT HANDLING

- Preserve every transcript utterance, including its ID, text, speaker role,
  and timestamps.
- Do not correct, paraphrase, delete, merge, or split utterances.
- You may attach existing utterance IDs to an event when their intervals
  overlap or the text clearly refers to that event.
- Do not create chapters or events from transcript utterances.

OUTPUT

Return only valid JSON conforming to the supplied schema.
\end{verbatim}
\tcbline
\begin{verbatim}
{
  "chapters": [
    {
      "chapter_id": "chapter_001",
      "start_time": 0.0,
      "end_time": 30.0,
      "events": [
        {
          "event_id": "chapter_001_event_001",
          "start_time": 4.0,
          "end_time": 8.0,
          "event_label": "scene_interaction_with_suspects_or_witnesses",
          "actor_role": "officer",
          "target_role": "member_of_public",
          "description": "An officer speaks to a member of the public.",
          "related_utterance_ids": ["utterance_003"]
        }
      ]
    }
  ],
  "utterances": []
}
\end{verbatim}
\end{promptbox}

\begin{promptbox}{Speech and speaker-attribution prompt and output schema (omni-modal base systems)}
\begin{verbatim}
Transcribe the audible speech in the recording.

For each utterance:
- provide start and end time;
- provide the transcript exactly as heard where possible;
- assign one speaker role: officer, member_of_public, other, or unknown;
- mark unintelligible words explicitly;
- mark overlapping speech;
- do not reconstruct missing words from context;
- do not infer speaker identity from appearance;
- do not assign a speaker role when the evidence is insufficient.

Preserve legally or evidentially important words such as negations, numbers, names,
commands, and threats. If a word is uncertain, use [unclear] rather than selecting the most
plausible word.
\end{verbatim}
\tcbline
\begin{verbatim}
{
  "utterances": [
    {
      "utterance_id": "u1",
      "start_time": 34.2,
      "end_time": 36.8,
      "speaker_role": "officer",
      "text": "Please put your hands where I can see them.",
      "speaker_confidence": "medium",
      "transcription_status": "supported",
      "overlap": false,
      "uncertain_tokens": []
    }
  ]
}
\end{verbatim}
\end{promptbox}

The judge returns a single binary label, so partially supported, uncertain and duplicate events are counted as unsupported, matching the conservative rule used for the human labels (Section~\ref{sec:judge}).
\begin{promptbox}{Judge prompt and output schema}
\begin{verbatim}
You are assessing whether a predicted event is supported by the supplied recording.

You are given:
- the recording;
- one predicted event;
- its time interval;
- its actor and target roles and description.

Assess only whether the event and its essential details are supported by information that is
visibly or audibly present in the recording. Do not use outside knowledge or infer legality,
intent, guilt, identity, or causation.

Return exactly one label and nothing else:

SUPPORTED
Use SUPPORTED only when the event and all material details, including the action, actor,
target, and timing, are directly observable or clearly audible in the recording.

UNSUPPORTED
Use UNSUPPORTED when the event or any material detail is absent, contradicted, materially
mistimed, overly specific, ambiguous, obstructed, inaudible, edited out, or not directly
established by the recording. Treat partially supported, uncertain, duplicate, and
unverifiable events as UNSUPPORTED.

Output exactly one of:
SUPPORTED
UNSUPPORTED
\end{verbatim}
\tcbline
\begin{verbatim}
{
  "type": "string",
  "enum": ["SUPPORTED", "UNSUPPORTED"]
}
\end{verbatim}
\end{promptbox}

\begin{promptbox}{Human coding form}
\begin{verbatim}
Review the predicted event against the recording.

Record:

1. Is the event directly supported?
   - supported
   - partially supported
   - unsupported
   - uncertain

2. Which event type is observed?
   - one of the 16 event types
   - none of the 16 types
   - cannot determine

3. Is the event interval adequate?
   - adequate
   - too early
   - too late
   - too broad
   - too narrow
   - cannot determine

4. Is the actor correct?
   - correct
   - incorrect
   - unknown

5. Is the target correct?
   - correct
   - incorrect
   - unknown

6. Does the description contain unsupported detail?
   - no
   - yes

7. What is the main reason for disagreement?
   - action
   - actor
   - target
   - timing
   - wording/detail
   - audio
   - visibility
   - editing/incomplete footage
   - other
\end{verbatim}
\end{promptbox}

\section{Datasheet for \BWVP{}}
\label{app:datasheet}
We follow the datasheet questions of \citet{gebru2021datasheets}.

\paragraph{Motivation.} \BWVP{} was created to test whether current video AI can support review of body-worn video, and to measure whether its outputs are traceable to the footage, how visible their failures are, and whether they can be contested (Section~\ref{sec:argument}). It was built by the authors at the Centre for Vision, Speech and Signal Processing, University of Surrey.

\paragraph{Composition.} The dataset has 27 recordings (uk01 to uk27) involving 16 UK police forces, 118 minutes in total, uploaded between 2017 and 2026; 3 were published by police forces and 24 by news outlets (Tables~\ref{tab:bwv-composition} and~\ref{tab:bwv-stats}). Each recording has a human-corrected transcript with speaker roles and timestamps, source-segment labels marking the retained body-worn footage, and 53 annotated events labelled with 16 event types, plus a video-level tag for force later challenged in court or at a hearing or inquest (Table~\ref{tab:inventory}). The judge validation adds 100 predicted events with blinded human support labels (Appendix~\ref{app:judge}). Recordings show identifiable people, including minors, and violent incidents, injuries and a death after restraint; they include footage shown in court, at an inquest and at misconduct hearings. No demographic attributes are recorded and no person is named. The dataset is a small pilot benchmark, not a representative sample of UK body-worn video, and has no training split.

\paragraph{Collection process.} Recordings were selected from footage already publicly released by police forces and news outlets, using the criteria in Section~\ref{sec:bwv-data}: wholly or mainly body-worn footage, at least one interaction between an officer and a member of the public, and variation in lighting, camera motion, occlusion, audio quality and number of people. No new data were collected from people, the people shown did not consent to this analysis, and no secure, proprietary or police-held footage was used.

\paragraph{Preprocessing and labelling.} Initial event annotations were generated by Baichuan-Omni-1.5 and MiMo-V2.5 and initial transcripts by Whisper (small.en); two annotators then reviewed and corrected them against the video and audio, following the guidelines in Appendix~\ref{app:guidelines}. Before annotation and evaluation, segments not filmed by a body-worn camera (such as studio, CCTV or helicopter inserts) and segments carrying broadcaster overlays were removed, so only clean BWV was annotated and evaluated; the original video files are not redistributed.

\paragraph{Uses.} The dataset is intended for evaluating how well video and audio models describe police-relevant footage, under human review. It should not be used to identify individuals, to infer demographic or personal attributes, to train surveillance systems, or to support decisions about the people shown. Because it is small and drawn from publicised incidents, results on it should not be read as estimates of performance on operational footage.

\paragraph{Distribution.} We release the source URL of each recording, the event annotations, the transcripts with speaker roles, the source-segment labels, and the judge and human support labels; the release link will be added to a later version of this preprint. No video files are redistributed: each recording remains available only from its original publisher, so a recording that is withdrawn by its publisher drops out of the benchmark.

\paragraph{Maintenance.} The authors maintain the annotations. A recording withdrawn by its publisher is removed at the next release, with the date of withdrawal recorded and affected results recomputed and versioned. Corrections and requests for removal can be sent to the corresponding author (armin.mustafa@surrey.ac.uk).

Table~\ref{tab:bwv-composition} lists the recordings, numbered uk01 to uk27 as in the event-type inventory (Table~\ref{tab:inventory}). Durations range from 26\,s (uk10, a short-form release) to 15\,min 44\,s (uk23), so recordings differ in how much of each incident they show and whether they were edited before release; several were news broadcasts combining body-worn footage with other material, which was removed before evaluation; lengths refer to the retained body-worn footage. Incidents are described by type only: people shown in the footage are not named, and source URLs are released with the annotations rather than printed here.

\begin{table*}[h]
\caption{Composition of \BWVP{}: 27 recordings involving 16 UK police forces (118 minutes). Source is the publishing channel; Year is the upload year; Length is min:s. BTP: British Transport Police.}
\label{tab:bwv-composition}
\footnotesize
\begin{tabular}{p{0.05\linewidth}p{0.17\linewidth}p{0.37\linewidth}p{0.15\linewidth}p{0.05\linewidth}p{0.06\linewidth}}
\toprule
ID & Force & Incident (de-identified) & Source & Year & Length \\
\midrule
uk01 & Metropolitan Police & Arrest after a stabbing & Guardian News & 2026 & 0:54 \\
uk02 & Metropolitan Police & Armed response vehicle pursuits (compilation, partly body-worn) & The Independent & 2025 & 10:10 \\
uk03 & Metropolitan / Northamptonshire & Arrest of a woman; footage initially said not to exist & BBC London & 2026 & 8:38 \\
uk04 & Merseyside & Arrest of the perpetrator of a mass stabbing & Manchester Evening News & 2025 & 1:50 \\
uk05 & Hampshire & Handcuffing of a man at a stabbing scene & The Independent & 2026 & 2:44 \\
uk06 & West Midlands & Officers speak to a man at a murder scene & Daily Record & 2026 & 6:26 \\
uk07 & Greater Manchester & Arrest at an airport (as shown in court) & LBC & 2025 & 4:59 \\
uk08 & Avon and Somerset & Alleged sledgehammer attack on an officer & Sky News & 2025 & 2:03 \\
uk09 & Avon and Somerset & Arrest after a street altercation (full recording) & ITV News & 2018 & 14:53 \\
uk10 & South Yorkshire & Arrests after an attack on a hotel during disorder & LBC & 2024 & 0:26 \\
uk11 & Northamptonshire & Pursuit of a lorry driver near a trunk road & Police (official) & 2023 & 1:41 \\
uk12 & Sussex & Man attacks officers with a hammer & PoliceWatch & 2017 & 0:59 \\
uk13 & Hampshire & Detention at a stabbing scene (news broadcast) & GB News & 2026 & 12:55 \\
uk14 & Metropolitan Police & Arrest after a stabbing in north London & DawnNews English & 2026 & 2:08 \\
uk15 & Metropolitan Police & Motorcycle officer confronts a man with a knife on an e-bike & Police (official) & 2026 & 7:13 \\
uk16 & Metropolitan Police & Events on the night of a fatal police shooting & The Mirror & 2024 & 2:33 \\
uk17 & Cheshire & Arrest of a suspect in a hospital murder investigation & TalkTV & 2023 & 0:44 \\
uk18 & Sussex / Metropolitan & Arrest of a couple after a search for a missing baby & The Telegraph & 2024 & 2:00 \\
uk19 & Merseyside & Arrest of a murder suspect & Times Radio & 2023 & 2:26 \\
uk20 & Cheshire & Arrests of two teenagers after a murder & Guardian News & 2024 & 1:23 \\
uk21 & Metropolitan Police & Vehicle stop later examined at a misconduct hearing & London Live & 2023 & 2:07 \\
uk22 & Essex & Arrest after a fatal stabbing at a constituency surgery & Sky News & 2022 & 2:22 \\
uk23 & Bedfordshire & Arrest after a triple murder & The Mirror & 2025 & 15:44 \\
uk24 & City of London & Taser use at a roadside stop & Guardian News & 2022 & 2:54 \\
uk25 & Cambridgeshire / BTP & Taser use and arrest after a train attack & The Sun & 2025 & 2:48 \\
uk26 & Dorset & Arrest of a 15-year-old, shown at a misconduct hearing & Police (official) & 2025 & 1:58 \\
uk27 & Metropolitan Police & Restraint of a man who later died, shown at an inquest & Sky News & 2020 & 3:02 \\
\bottomrule
\end{tabular}
\end{table*}

\begin{table}[h]
\caption{\BWVP{} statistics.}
\label{tab:bwv-stats}
\begin{tabular}{lr}
\toprule
 & \BWVP{} \\
\midrule
Videos & 27 \\
Police forces & 16 \\
Published by police / news outlets & 3 / 24 \\
Upload years & 2017--2026 \\
Duration range & 0:26 to 15:44 \\
Total duration (min) & 118 \\
Mean / median duration (min:s) & 4:22 / 2:26 \\
Event types (scored / video-level tag) & 16 / 1 \\
Recordings with contested-force tag & 6 \\
Annotated events & 53 \\
\bottomrule
\end{tabular}
\end{table}

\section{KnifeHunter Training, Backbones and Ablation}
\label{app:kh-ablation}
\paragraph{KHNet architecture.} Each model is a backbone (EfficientNet-B7 \cite{tan2019efficientnet}, ConvNeXt-Base \cite{liu2022convnext}, SwinV2-Base \cite{liu2022swinv2} or EVA02-Base \cite{fang2024eva02}) producing a feature tensor $X\in\mathbb{R}^{W\times H\times K}$, followed by an aggregation head: GeMNet \cite{radenovic2019gem}, ActNet \cite{husain2021actnet}, TokenNet \cite{wu2022tokennet}, SENet \cite{lee2023senet} or our KHNet. KHNet has two branches. The local branch applies regional $L_p$ pooling \cite{shao2023superglobal} over an $s\times s$ neighbourhood $\mathcal{N}(w,h)$ and averages it with the input,
\begin{equation}
R_{w,h,k}=\Big(\frac{1}{s^2}\sum_{(i,j)\in\mathcal{N}(w,h)}|X_{i,j,k}|^{p_r}\Big)^{1/p_r},\qquad G_{w,h,k}=\tfrac{1}{2}\big(X_{w,h,k}+R_{w,h,k}\big).
\end{equation}
The global branch applies a $3\times3$ convolution, batch normalisation, SiLU, a learnable Weibull activation, global average pooling and power normalisation \cite{husain2021actnet}, giving $f^{(w)}$. Local features are fused by projecting them onto the orthogonal complement of the global descriptor \cite{yang2021dolg},
\begin{equation}
f^{(\mathrm{orth})}_{w,h}=G_{w,h}-\frac{G_{w,h}^{\top}f^{(w)}}{\lVert f^{(w)}\rVert^{2}}\,f^{(w)},
\end{equation}
concatenating them with the broadcast global vector and average pooling. A projection head (linear layer, batch normalisation, PReLU) gives a 512-dimensional descriptor. Training settings follow.

\paragraph{Training.} Images are resized to $448\times448$. Models are trained end to end for 25 epochs as 543-way classifiers with sub-center ArcFace (margin 0.3, scale 45), Adam (learning rate $10^{-5}$, decayed by 0.8 every 4 epochs) and batch size 128. The GeM exponent is initialised at $p=3$, the Weibull parameters at $\alpha=100$, $\beta=2$, $\gamma=100$, $\zeta=2$, and $p_r=2.5$.

\begin{table}[h]
\caption{Backbone comparison with GeMNet on \KH{} (mAP, \%).}
\label{tab:kh-backbones}
\footnotesize
\begin{tabular}{lrrrrr}
\toprule
Backbone & Medium & Hard & Medium+D & Hard+D & Params \\
\midrule
EfficientNet-B7 \cite{tan2019efficientnet} & 79.4 & 56.1 & 73.6 & 45.5 & 70M \\
ConvNeXt-Base \cite{liu2022convnext} & 81.2 & 58.1 & 79.1 & 53.8 & 88M \\
SwinV2-Base \cite{liu2022swinv2} & 82.4 & 59.2 & 80.2 & 54.4 & 88M \\
EVA02-Base \cite{fang2024eva02} & 83.1 & 61.2 & 81.5 & 57.9 & 88M \\
\bottomrule
\end{tabular}
\end{table}

\begin{table}[h]
\caption{Cumulative contribution of KHNet components on EVA02-Base (mAP, \%). Each row adds to the row above.}
\label{tab:kh-ablation}
\footnotesize
\begin{tabular}{lrrrr}
\toprule
Configuration & Medium & Hard & Medium+D & Hard+D \\
\midrule
Global average pooling & 81.1 & 58.2 & 79.1 & 53.1 \\
GeMNet & 83.1 & 61.2 & 81.5 & 57.9 \\
+ Weibull pooling (ActNet) & 83.9 & 62.1 & 82.6 & 58.9 \\
+ Local branch, orthogonal fusion & 85.6 & 64.1 & 84.7 & 62.1 \\
+ Regional $L_p$ refinement (KHNet) & 86.5 & 65.3 & 85.7 & 63.4 \\
\bottomrule
\end{tabular}
\end{table}

\end{document}